\pdfoutput=1

\documentclass[11pt]{article}

\usepackage[final]{acl}
\usepackage{times}
\usepackage{hyperref}
\hypersetup{breaklinks=true}
\usepackage{latexsym}
\usepackage{amsmath}
\usepackage{multirow}
\usepackage[T1]{fontenc}
\usepackage{booktabs}

\usepackage[utf8]{inputenc}

\usepackage{microtype}
\usepackage{bbm}
\usepackage{inconsolata}

\usepackage{graphicx}

\definecolor{pos}{RGB}{167, 199, 231}
\definecolor{neg}{RGB}{250, 160, 160}
\newcommand{\clerc}{\textsc{Clerc}}
\newcommand{\clercs}{\textsc{Clerc\ }}
\newcommand{\clercp}{\texttt{CLERC/passage}}
\newcommand{\clercd}{\texttt{CLERC/doc}}
\newcommand{\clercg}
{\texttt{CLERC/generation}}

\definecolor{darkblue}{rgb}{0, 0, 0.5}
\definecolor{ballblue}{rgb}{0.13, 0.67, 0.8}
\definecolor{crimson}{rgb}{0.7529,0.2235,0.1686}
\definecolor{darkgreen}{rgb}{0.0, 0.6, 0.2}
\definecolor{red4cb}{rgb}{0.8353,0.2431,0.2431}
\definecolor{blue4cb}{rgb}{0.1804,0.4000,0.6745}
\hypersetup{colorlinks=true, citecolor=darkblue, linkcolor=darkblue, urlcolor=darkblue}

\newcommand{\goodtext}[1]{\textcolor{darkgreen}{\textit{{#1}}}}

\newcommand{\misstext}[1]{\textcolor{brown}{\texttt{{#1}}}}

\newcommand{\badtext}[1]{\textcolor{red4cb}{\underline{{#1}}}}

\title{\clerc: A Large Long-Context Dataset for Retrieval and Reasoning}
\title{\clerc: A Long-Context Dataset for Legal Citation Retrieval \\ and Retrieval-Augmented Case Generation}
\title{\clerc: A Long-Context Dataset for Legal Citation Retrieval \\ and Retrieval-Augmented Analysis Generation}
\title{\clerc: A Dataset for Legal Case Retrieval \\ and Retrieval-Augmented Analysis Generation }

\author{
  Abe Bohan Hou\textsuperscript{$\clubsuit$} \quad Orion Weller\textsuperscript{$\clubsuit$} \quad Guanghui Qin\textsuperscript{$\clubsuit$} \quad \\
  \textbf{Eugene Yang}\textsuperscript{$\clubsuit$} \quad \textbf{Dawn Lawrie}\textsuperscript{$\clubsuit$} \quad \textbf{Nils Holzenberger}\textsuperscript{$\spadesuit$} \\
  \textbf{Andrew Blair-Stanek}\textsuperscript{$\diamondsuit$}\textsuperscript{$\clubsuit$} \quad \textbf{Benjamin Van Durme}\textsuperscript{$\clubsuit$} \\
  \textsuperscript{$\clubsuit$} Johns Hopkins University \quad   \\ \textsuperscript{$\spadesuit$}Télécom Paris, Institut Polytechnique de Paris \\
\textsuperscript{$\diamondsuit$} University of Maryland, Carey School of Law
}

\newcommand{\dataset}{\textsc{\clerc}}

\newcommand{\floor}[1]{\lfloor {#1} \rfloor}

\begin{document}

\maketitle

\begin{abstract}
  Legal professionals need to write analyses that rely on citations to relevant precedents, i.e., previous case decisions.  
  Intelligence systems assisting legal professionals in writing such documents provide great benefits but are challenging to design.
  Such systems need to help locate, summarize, and reason over salient precedents in order to be useful.
  To enable systems for such tasks, we work with legal professionals to transform a large open-source legal corpus into a dataset\footnote{Code/data is publicly available at \url{https://github.com/bohanhou14/CLERC} and \url{https://huggingface.co/datasets/jhu-clsp/CLERC}} supporting two important backbone tasks: information retrieval (IR) and retrieval-augmented generation (RAG). 
  This dataset \dataset\ (\textbf{C}ase
\textbf{L}aw \textbf{E}valuation and \textbf{R}etrieval \textbf{C}orpus), is constructed for training and evaluating models on their ability to (1) find corresponding citations for a given piece of legal analysis and to (2) compile the text of these citations (as well as previous context) into a cogent analysis that supports a reasoning goal.
  We benchmark state-of-the-art models on \dataset, showing that current approaches still struggle: GPT-4o generates analyses with the highest ROUGE F-scores but hallucinates the most, while zero-shot IR models only achieve 48.3\% recall@1000. 

\end{abstract}

\section{Introduction}
Writing legal analyses is a central task of lawyers. For example, in the U.S. legal system,  litigation attorneys draft motions and briefs, citing relevant cases (case law) to support their analysis. From a corpus of millions of cases, finding the relevant files and weaving them into a persuasive whole requires substantial time and effort. Enhancing lawyers' capacity to retrieve relevant cases and generate legal analyses presents difficult challenges across various fields, including information retrieval, natural language processing, and the broader AI community.
\begin{figure*}
\centering
\includegraphics[scale=0.65]{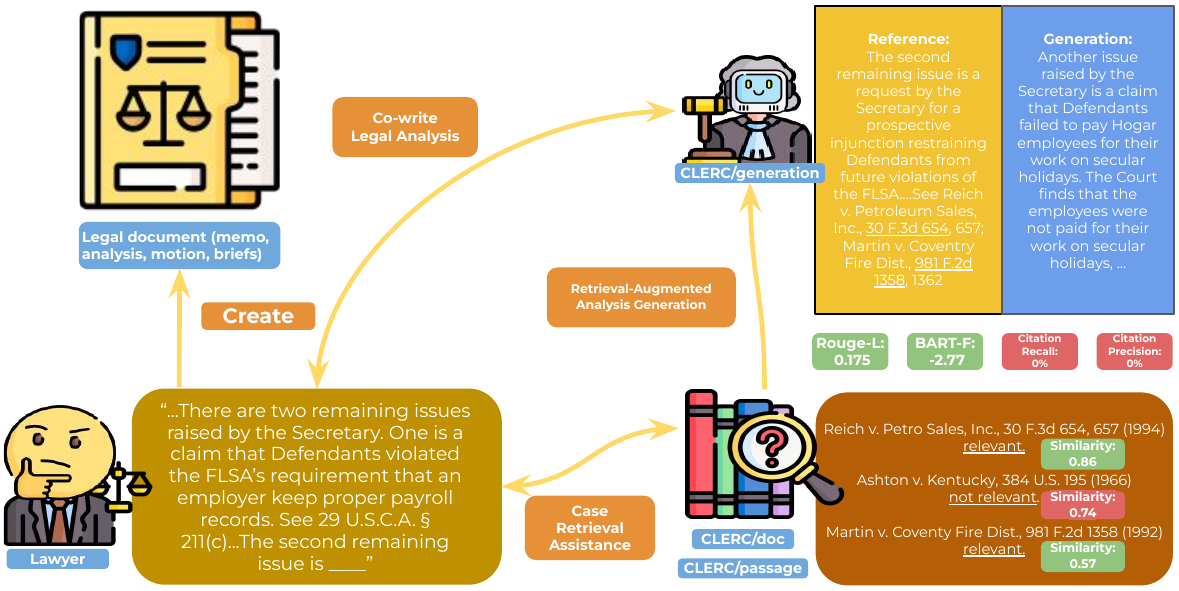}%
\caption{An overview of how \clercs enables systems that empower legal professionals: retrieval models for \clercd\ and \clercp\ that assist with finding relevant cases to support the analysis, and retrieval-augmented generation systems with \clercg\ that aid in legal analysis generation.}
\label{fig:pipeline}
\end{figure*}

To support the development of information retrieval and generative language models in writing legal documents, we build a dataset based on existing case law, i.e., the \textit{precedents}. Case law is written by judges to explain the court's reasoning for its resolution of legal disputes based on concrete facts and laws. It is foundational for legal analyses for Common Law and some Civil Law jurisdictions \citep{ma2021lecard, mahari2023LePaRD}.
Curating a quality dataset based on case law is valuable for two main reasons.
First, given that case law is carefully written and serves as a strong prediction of future case law, the data is valuable for developing legal intelligence systems. Second, given the desire for retrieval accuracy and text generation quality, high-quality training data is critical for producing effective models.

Thus, we build a dataset, \clercs (\textbf{C}ase
\textbf{L}aw \textbf{E}valuation and \textbf{R}etrieval \textbf{C}orpus), as a step towards enabling collaborative AI systems that assist legal professionals, while also providing AI researchers with a challenging and real-world evaluation set. \clercs is built upon digitized case law from the Caselaw Access Project (CAP)\footnote{\url{https://case.law/}} \citep{cap} by Harvard Law School, which contains over 1.84 million federal case documents with an average of 11.54 citations per document (see a taxonomy of U.S. legal data in Figure \ref{fig:data-taxonomy}). We transform this corpus into a comprehensive task dataset that can be used for information retrieval (IR) and retrieval-augmented generation (RAG). Compared with existing precedent datasets \citep{hendersonkrass2022pileoflaw, santosh24ecthr, mahari2023LePaRD, ma2021lecard, guha2023legalbench}, we offer the following novel contributions: 
\begin{enumerate}
\item Through working with legal professionals, we provide a formulation of legal case retrieval and generation tasks that balance the needs and perspectives of legal professionals with computational feasibility.

\item We build an open-source pipeline for transforming CAP into a large-scale, high-quality dataset designed for training and evaluating models on legal IR and RAG tasks.
    
\item We conduct a comprehensive evaluation of long-context case retrieval and retrieval-augmented legal analysis generation on \dataset{} with state-of-the-art methods, revealing that IR models struggle to retrieve relevant documents and LLMs frequently hallucinate.
\vspace{-2mm}
\end{enumerate}

\begin{figure*}[t]
    \centering
    \fbox{
    \begin{minipage}{38.5em}
    \footnotesize
    \textbf{Query: }
    \dots Summary judgment should be granted where ``the pleadings, depositions, answers to interrogatories and admissions on file, together with the affidavits, if any, show there is no genuine issue as to any material fact and that the moving party is entitled to judgment as a matter of law.'' \textcolor{cyan}{Fed.R.Civ.P. 56(c)}. The moving party has the responsibility of informing the Court of portions of the record or affidavits that demonstrate the absence of a triable issue. \textcolor{violet}{\underline{Celotex Corp. v. Catrett, 477 U.S. 317, 322, 106 S.Ct. 2548, 91 L.Ed.2d 265 (1986).}} The moving party may meet its burden of showing an absence of disputed material facts by demonstrating \textcolor{brown}{\textit{``that there is an absence of evidence to support the non-moving party’s case.''}} \textcolor{purple}{\texttt{Id. at 325, 106 S.Ct. 2548}}. Any doubt as to the existence of a genuine issue for trial is resolved against the moving party. \textcolor{purple}{\texttt{Anderson v. Liberty Lobby, Inc., 477 U.S. 242, 255, 106 S.Ct. 2505, 91 L.Ed.2d 202 (1986)}}; \dots
    
    \end{minipage}
    }
    \caption{A depiction of a query from \dataset{}. ``\textcolor{cyan}{Fed.R.Civ.P. 56(c)}'' cites a federal statute. \textcolor{violet}{\underline{Underlined violet}} is the query's \textbf{central citation sentence}. \textcolor{brown}{\textit{The brown italicized sentence}} is a \textbf{direct quote} from Celotex Corp. v. Catrett, making this query a \emph{direct query}. \textcolor{purple}{\texttt{The red citations in teletype fonts}} are \emph{non-central citations}. We define \texttt{single-removed} as a data view which masks only the central citation and its salient sentence, while \texttt{all-removed} masks all citations from the passage. All references to statutes are retained.}
    \vspace{-4mm}
    \label{fig:egquery}

\end{figure*}

\section{Related Work}

In this section, we provide an overview of long-context retrieval and RAG, including existing applications in the legal domain. 

\subsection{Legal Case Retrieval and Generation}
Case law retrieval is a longstanding problem aimed at helping legal experts find cases to cite \citep{wilson1962computer}. Techniques range from lexical search \citep{blair1985evaluation} to more recent semantic similarity-based approaches \citep{ebietomere2019semantic,askari2021combining}. Overall, most legal retrieval is done through WestLaw\footnote{\url{https://legal.thomsonreuters.com/en/westlaw}} or LexisNexis,\footnote{\url{https://www.lexisnexis.com/}} or services offered by Thomson Reuters,\footnote{\url{https://legal.thomsonreuters.com/en/products/court-express}} which charge large sums for access to a manually annotated and custom-based ontology. Although many new datasets exist for countries outside of the United States \citep{ma2021lecard, li2023lecardv2, santosh24ecthr, goebel2023coliee}, few existed in the United States before the large U.S. case law resource, CAP, was released to the public in 2023 \citep{Mahari2021AutoLAWAL}. A summary of more related work of legal IR can also be found in \citet{locke2022case}. Due to the recent advances of Large Language Models (LLMs) and RAG, legal generation is a new field with preliminary works, including analysis generation using small datasets or small-scale models \citep{tuvey2023argument, Li2021CourtOG, Lin2024LegalDD}, generating stories to educate about legal concepts \citep{Jiang2024LeveragingLL}, generating legal summaries \citep{Gesnouin2024LLaMandementLL, Shen2022MultiLexSumRS}, and profiling hallucinations in legal generation \citep{Dahl2024LargeLF, Magesh2024HallucinationFreeAT}. In comparison, we provide an extensive benchmark on \textbf{retrieving citations and generating long-form legal analysis}, which is a core task for lawyers that requires substantial time and effort.

\subsection{Long-Context Generation}

LLMs are limited in context length due to concerns of efficiency and performance ~\citep{longchat23}.  
Benchmarks have been introduced to measure the effectiveness of LLMs, including \citet{shaham-etal-2022-scrolls,tay2020LongRangeArena,dong2024BAMBOOComprehensiveBenchmark}.
Various methods were proposed to solve this problem, such as efficient attention~\citep[inter alia]{longformer20,performer21,qin-etal-dodo-2024}, recurrence~\citep{dai-etal-2019-transformer,rmt22}, context length generalization~\citep{roformer24,chen2023ExtendingContextWindow}, and RAG~\citep{rag20},
while some recent work challenges the effectiveness of those methods~\citep{lost24,qin-etal-2023-nlp}.
Recent proprietary LLMs are reported to extend the context length up to 1 million tokens, such as Gemini 1.5~\citep{gemini1.524}, Claude 3~\citep{claude324}, and GPT-4~\citep{gpt424}.
More specific to legal texts, generative models on long contexts are commonly used for legal case outcome prediction~\citep{qin2024ExplicitlyIntegratingJudgment,cao2024PILOTLegalCase}, but long-context generation datasets on legal cases are largely absent from the literature.

\subsection{Long-Document Retrieval}
Many state-of-the-art neural retrieval models, because of their reliance on BERT models \citep{devlin2018bert}, have a context limit of 512 tokens \citep{khattab2020colbert, santhanam2022colbertv2, ren2021rocketqav2, karpukhin-etal-2020-dense}. Accompanied by the advances in long-context LLMs \citep{brown2020language, touvron2023llama, gpt424}, there are also rising interests in long-context retrieval models \citep{fu2023monarch, gunther-etal-2023-jina, wang2023improving}. However, existing benchmarks lack insights on how both BERT-based and long-context models perform on well-defined legal retrieval tasks \citep{thakur2021beir, saad-falcon2024long, Shen2022MultiLexSumRS}.

\section{Dataset Construction}
 
\subsection{Dataset Preprocessing}

 \clercs contains a collection of long case documents (\clercd) and chunked documents for retrieval (\clercp) as well as for generation (\clercg). 
 We obtain the collection of raw case documents from CAP \citep{cap}, including all federal case law in U.S. history up to September 21, 2021. To prepare \clercd, we concatenate all types of opinions (e.g., majority, dissenting, concurring opinions) in every case document and remove the newline characters. To prepare \clercp, we split documents into chunks of 350 words, with a sliding window of 175 words. We discuss the \clercg\ preparation process in Section \ref{sec:arg-gen} in more detail.

Our final dataset consists of 1.84M documents with over 20.7M total citations, 23.7M passages for retrieval, and 6k passages for generation. Details and summary statistics are presented in Table \ref{tab:statistics}.

\subsection{Case Retrieval}
\label{sec:dataset-retrieval}

\paragraph{Query Definition}

As lawyers write legal analyses, they need to find case documents to support their texts. To simulate this use case, we define a query as a block of text, taken from a case document, with its \underline{citation in the middle} (a.k.a the central citation) \underline{removed}. The \emph{relevant passage} of a query is defined as the passage its central citation cites to. We provide both left and right contexts, as legal analysis involves context from both sides. 

\paragraph{Direct and Indirect Queries} A case document may cite another case in the form of a direct quote. \citep{bluebook}. We define two kinds of legal queries based on whether the central citation contains a direct quote (\texttt{direct}) or not (\texttt{indirect}). Ideally, using \textit{only} this direct quote to retrieve should yield an almost perfect accuracy through an exact string match. We build a heuristic program to recognize such direct quotes (detailed in Appendix \ref{app:parse}), and classify queries into direct and indirect queries.
We retrieve with \textit{only} direct quotes in Table \ref{tab:dq-res} and Section \ref{app:dq}.

\paragraph{Data Views} 
We define two data views of legal queries depending on whether all (\texttt{all-removed}) or only the central (\texttt{single-removed}) citation information is removed. Since each kind of query has two data views, this leads to a total of four categories of queries, illustrated in Figure \ref{fig:egquery}.

\begin{figure}
    \centering
    \resizebox{\columnwidth}{!}{%
    \includegraphics{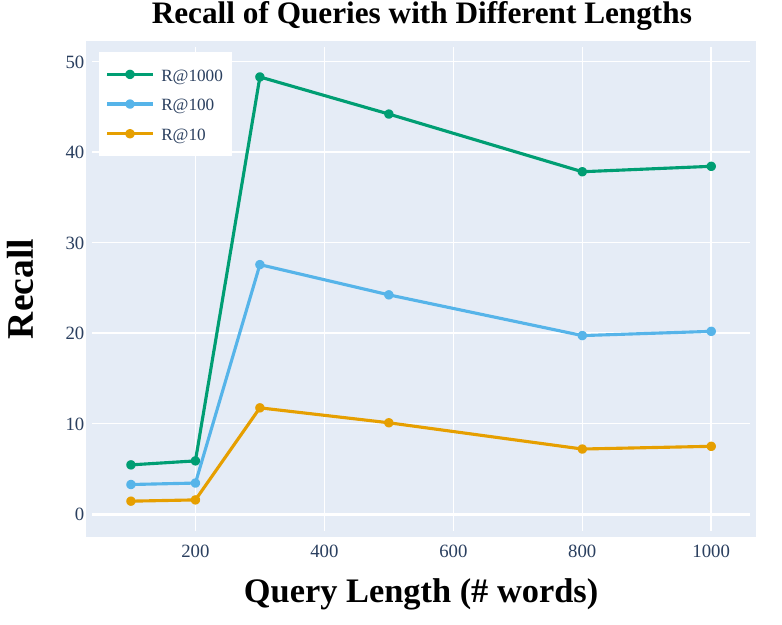}%
    }
    \caption{Retrieval results of BM25 with \clercs queries of length 100 to 1000 words. R@X represents Recall@X. \textbf{Queries with 300 words maximize recall and are used for our main experiments.}}
    \label{fig:abla-lengh}
     \vspace{-4mm}
\end{figure}

\paragraph{Query Construction}
First, we identify case citations in documents. For every case citation, we expand a window of 300 words centered on this citation to construct a query, as legal analysis uses context from both sides. We experimented with various query lengths and found that including 300 words yields the highest recall
(summarized in Figure \ref{fig:abla-lengh}), which indicates that this length provides enough critical information for retrieving relevant cases without distracting the models to retrieve other cases, as we discuss in Section \ref{exp:ret-abla}. We remove the central citation in this paragraph for \texttt{single-removed} queries and all case citations in this paragraph for \texttt{all-removed} queries. 

\paragraph{Query Extraction} To identify the case citation in a query, we use \texttt{eyecite} \citep{eyecite}, a Regex-based parser that recognizes case citations from the CAP corpus. We develop a heuristic and rule-based Python program to detect the sentence boundary where the case citation is located. We create the heuristic rules based on the \textit{Bluebook} \citep{bluebook} (detailed in \ref{app:parse}), which is the standard citation guide for U.S. legal professionals. We manually evaluate the extraction accuracy on 1000 sentences and discovered that 87.8\% of citation sentences can be accurately identified and removed.

\begin{figure}
    \centering
    \resizebox{\columnwidth}{!}{%
    \includegraphics{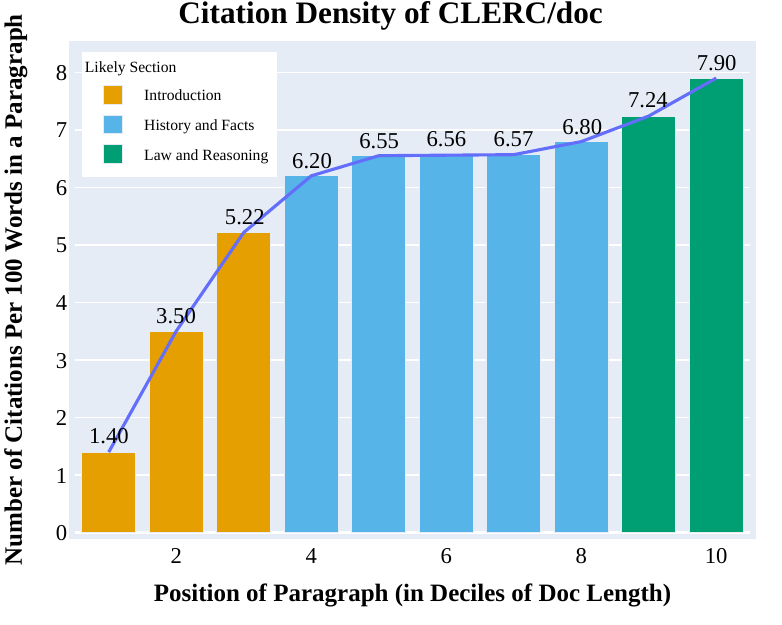}%
    }
    \caption{Number of citations per 100 words in a paragraph, arranged by the positions (in deciles) of paragraphs in the document. \textbf{Documents tend to cite more often in later paragraphs.} Case law typically has different sections (e.g. Intro, Facts, Reasoning, etc). Deciles with the same color code likely belong to the same general sections of case law..
    }
    \label{fig:citation_density}
    \vspace{-4mm}
\end{figure}
\subsection{Legal Analysis Generation}
\label{sec:arg-gen}

We prepare \clercg\ to assess the ability of LLMs to generate realistic legal analyses from previously written text and given references. To evaluate the generation quality, we feed in the preceding contexts of a target analytical paragraph (reference) and other relevant cases cited by the paragraph, prompt LLMs to generate a new analytical paragraph, and compare the generation with the reference from the original document. 

\paragraph{Analytical Paragraphs Identification} We observe that case documents usually start with introducing the basic facts about the cases, and then make multiple legal analyses while citing other relevant cases (see Figure \ref{fig:egdoc}) \citep{lippmann_criminal_procedure}. To prepare the reference for evaluating analysis generation, we only use paragraphs (blocks of text separated by newline characters) containing legal analyses and discard the rest.

Our rule of thumb to determine if a paragraph is a suitable reference is whether it has a large number of case citations, as the analytical claims are supported by case. 
In Figure \ref{fig:citation_density}, we plot the trend of citation density in case law documents (only the majority opinion), discovering that cases cite more often, which is consistent with our prior observation. 

\paragraph{Reference Selection}
We sample analytical paragraphs from case texts as reference for evaluating analysis generation.
As we observe that analysis tends to occur in later paragraphs of a case document, we randomly sample from the last 33.3\% of all paragraphs but exclude the last two paragraphs (which are usually short conclusions) and paragraphs with fewer than two case citations.

\paragraph{Input Preparation}
After we sample the reference paragraph, we take all texts preceding the reference as a part of the input. We concatenate the most salient passages of relevant cases cited by the reference paragraph, whose chunks are selected via RankLlaMa \citep{Ma2023FineTuningLF}.

\begin{table}[t!]
\small
\begin{tabular}{l|@{\hspace{0.6\tabcolsep}}c@{\hspace{0.9\tabcolsep}}c}
\toprule
Dataset Statistic & \# instances (M) & Avg len (words) \\
\midrule
\clercd & 1.84 & 2279 \\
\clercp & 23.7 & \phantom{0}350 \\
\texttt{CLERC/queries} & 0.105 & \phantom{0}300 \\
\midrule
\clerc & 25.5 & N/A \\
LePaRD & 0.375 & 200 char \\
ECtHR-PCR & 0.015 & 5530 \\
LeCaRDv2 & 0.055 & 4766 \\
\midrule
\clercg &0.006 &3371 \\
\bottomrule
\end{tabular}
\caption{Dataset statistics. \clercs has the largest document and passage combined collection compared to other legal retrieval datasets \citep{li2023lecardv2, mahari2023LePaRD, ecthr}. Additionally, \clercs contributes a novel subset for legal analysis generation.}
\label{tab:statistics}
\end{table}

\begin{table}[t]
\centering
\small
\begin{tabular}{@{}lrrrr@{\hspace{0.8\tabcolsep}}r@{}}
\toprule
                    Type
                    & \multicolumn{1}{l}{R@5}  & \multicolumn{1}{l}{R@10}  & R@100 & \multicolumn{1}{l}{R@1K}& nDCG@10  \\ \midrule
ISR & \textbf{7.9}                          & \textbf{11.8}                          & \textbf{27.6}       & 48.3                            & \textbf{5.4}                         \\
IAR   & 7.5                          & 11.1                          & 25.1       & 43.9                            & 5.2                         \\
DSR   & 7.0                          & 11.0                          & 26.7       & \textbf{48.5}                            & 5.0                         \\ 
DAR      & 6.8                          & 9.88                          & 24.9       & 46.3                            & 4.6                         \\ \bottomrule
\end{tabular}
\vspace{1ex}
\caption{Retrieval results of BM25 with different types of \clercs queries. I stands for \texttt{indirect}, D for \texttt{direct}, SR for \texttt{single-removed}, and AR for \texttt{all-removed}. E.g. ISR refers to \texttt{indirect single-removed}. All numbers are in percentage. R@X represents Recall@X. \textbf{ISR queries are the easiest queries to retrieve and are used for our main retrieval experiments.}}
\label{tab:retrieval-abla}
\vspace{-2mm}
\end{table}
\begin{table*}[t]
\centering
\small
\begin{tabular}{llllllrrr}
\toprule 
\multirow{2.5}{*}{Type} & \multirow{2.5}{*}{Params} & \multirow{2.5}{*}{Length} & \multirow{2.5}{*}{Model} & \multicolumn{4}{c}{Metrics} \\
\cmidrule{5-8}
 &  &  &  & R@10 $\uparrow$ & R@100 $\uparrow$ & R@1K $\uparrow$ & nDCG@10 $\uparrow$\\
\midrule 
Sparse & N/A  & N/A & BM25 &11.7  &27.8  &48.3 &5.40\\
\midrule
 \multirow{3}{*}{Late Interaction} 
 & 110M & 512 & ColBERTv2 &2.2  &7.1  & 17.6 &1.05  \\
 & 110M & 2048 &Jina-ColBERT-v1-en &2.1  &5.8  &14.2 &1.15\\
 & 110M & 4096 &Jina-ColBERT-v1-en &2.2  &6.9  &16.1 &1.19\\
 
\midrule
 \multirow{4}{*}{Bi-Encoders} 
  & 110M & 512 & DPR & 4.9 & 13.1 & 26.2 &  2.24           \\
 & 110M & 512 & BGE En v1.5 & 7.7 & 23.2 & 42.4 & 3.43 \\
 & 110M & 512 & E5 v2  & 8.4  & 24.3 & \underline{42.3} & 3.67 \\
 & 110M & 512 & Contriever-MSMarco & 9.3 & 24.8 & 41.4 & 4.22 \\
 \midrule

 \multirow{2}{*}{Fine-Tuned}   & 110M & 512 & ft-DPR  &18.6  &40.9  &63.1 &11.59    \\
 & 110M & 512 & ft-LegalBERT DPR &\textbf{23.2}  &\textbf{46.9}  &\textbf{68.5}  &\textbf{14.67}             \\
\bottomrule
\end{tabular}
\caption{Retrieval results on \clercd\ and \clercp, evaluated with Recall@10, @100, @1000 and nDCG@10.  Overall, zero-shot models are particularly bad due to distribution shifts between standard retrieval training data and the legal domain (Section~\ref{sec:discuss-ret}). Models fine-tuned on legal data show effectively improved results.
\label{tab:ret-main}
}
\end{table*}

\begin{table*}[]
\centering
\small
\begin{tabular}{@{}lr@{\hspace{0.9\tabcolsep}}r@{\hspace{0.9\tabcolsep}}r@{\hspace{0.9\tabcolsep}}rrrr@{}}
\toprule
Model                    & \multicolumn{1}{l}{R1 $\times 100$} $\uparrow$ & \multicolumn{1}{l}{R2$\times 100$} $\uparrow$ & \multicolumn{1}{l}{RL$\times 100$} $\uparrow$& \multicolumn{1}{l}{BF} $\uparrow$& \multicolumn{1}{l}{CR $\times 100$}$\uparrow$ & \multicolumn{1}{l}{CP$\times 100$} $\uparrow$ &\multicolumn{1}{l}{CFP$\times 100$} $\downarrow$ \\ \midrule
Mistral-7B-v0.3 & \underline{23.78} / 18.48          & \underline{8.00} / 4.63               & \underline{21.89} / 16.78          & \underline{-3.35} / -3.57          & \underline{42.67} / 0.32           &\underline{32.65} / 0.70    &\underline{5.27} / 25.28                    \\
gemma-1.1-7b-it          & \underline{18.33} / 16.82          & \underline{5.28} / 3.99            & \underline{16.70} / 14.81           & \underline{-3.71} / -3.62          & \underline{37.22} / 0.00              &\underline{36.59} / 0.00           &\textbf{\underline{4.34}} / 84.62             \\
GPT-4o                   & \underline{\textbf{26.82}} / 19.87          & \underline{\textbf{10.04}} / 5.56           & \underline{\textbf{24.89}} / 17.93          & \underline{-3.33} / -3.64          & \underline{\textbf{89.87}} / 3.28           &\textbf{\underline{52.84}} / 7.66  &\underline{6.41} / 71.49  \\
Llama-3 & \underline{25.16} / 18.68          & \underline{8.91} / 4.68            & \underline{23.42} / 16.86          & \underline{\textbf{-3.16}} / -3.48          & \underline{62.64} / 0.07           &\underline{33.41} / 0.84  &\underline{4.60} /~~ 7.17                      \\
\midrule
Avg Score &18.55 &4.73 &16.71 &-3.52 &0.84 &2.13 &40.64\\
~~~+ ref &23.50 &7.96 &21.70 &-3.38 &54.88 &36.39 &5.13 \\
Avg Gain (\%) &+ 26.63 &+ 68.08 &+ 29.87 &+ 3.98 &+ 6471.43 &+ 1603.76 & - 87.37 \\
\bottomrule
\end{tabular}
\caption{Analysis generation results on \clercg, evaluated with F-scores of ROUGE and BARTScore (BF), CR, CP, and CFP. All models are instruct versions, and Llama-3 is shortened from Meta-Llama-3-8B-Instruct. Left / right column results correspond to prompting with / without the texts of cited cases. \textbf{GPT-4o generates analyses with the highest ROUGE F-scores but hallucinates the most}. \textbf{Prompting with cited cases shows effective improvement across all metrics}, especially boosting CR and CP and lowering CFP, making more factual citations in generations. 
}
\label{tab:arg-gen-res}
\vspace{-2mm}
\end{table*}
\begin{figure*}[t]
    \centering
    \fbox{
    \begin{minipage}{40em}
    \footnotesize
    
FINDINGS OF FACT AND CONCLUSIONS OF LAW

PAEZ, District Judge.

\textbf{I. INTRODUCTION}

The Board of Trustees of the Aircondi-tioning and Refrigeration Industry Health and Welfare Trust Fund (“Health Fund”), the Board of Trustees of the Aircondition-ing and Refrigeration Industry Retirement Trust Fund (“Retirement Fund”), and the Board of Trustees of the Airconditioning and Refrigeration Industry Defined Contribution Retirement Plan (“401(k) Plan”) (collectively, the “Trust Funds”) brought this action against J.R.D. Mechanical Services, Inc. ...(Omitted for brevity)...\\

\textbf{II. FINDINGS OF FACT}

1. Defendant J.R.D. is a California corporation and at all material times was an employer within the meaning of section 3(5) of ERISA, 29 U.S.C. § 1002(5) ...(Omitted for brevity)...\\

\textbf{III. CONCLUSIONS OF LAW}

Based on the forgoing findings of fact, the Court makes the following conclusions of law:
...(Omitted for brevity)...\\

\textcolor{ballblue}{“‘Any’ control over disposition of plan money makes the person who has the control a fiduciary.” \textit{IT Corp. v. General Am. Life Ins. Co., 107 F.3d 1415, 1421 (9th Cir.1997), cert. denied, 522 U.S. 1068, 118 5.Ct. 738, 139 L.Ed.2d 675 (1998).} Thus, “a person with authority to direct payment of a plan’s money [is] deemed a fiduciary.” Id. Divers’ status as a corporate officer does not exempt him from liability. See \textit{Kayes v. Pacific Lumber Co., 51 F.3d 1449, 1459 (9th Cir.1995)} (“This court has held corporate officers to be liable as fiduciaries on the basis of their conduct and authority with respect to ERISA plans.”). Rather, he is liable because, as established by the factual findings, he exercised authority and control over the management and disposition of the Trust Funds’ assets.}
\\
...(Omitted for brevity)...
 \end{minipage}
} 
    \caption{Example case document (\textit{99 F. Supp. 2d 1115}). \textcolor{ballblue}{Blue texts} are analytical texts and \textit{italicized texts} are case citations. This is a typical case document which starts with introductions and facts (\textbf{I} and \textbf{II}) about the case and then dives into legal analyses and reasoning at later paragraphs (\textbf{III}). See casetext link for full text: \url{https://casetext.com/case/board-of-trustees-of-aircondition-v-jrd-mech}.}
    \vspace{-2mm}
    \label{fig:egdoc}
\end{figure*}

\section{Experiments}
In this section, we present experiment results of legal citation retrieval (Section \ref{exp:retrieval}) and legal analysis generation (Section \ref{exp:arg-gen}) with the proposed \clerc.

\subsection{Finding Legal Citations}\label{exp:retrieval}

\paragraph{Task Formulation}
Given a query paragraph $q$ with a central citation sentence $s$ with left and right contexts $l$ and $r$   ($q = l \circ s \circ r$, where $\circ$ indicates string concatenation), retrieve the relevant cited passages or documents $R = \{d_a, ..., d_b\}$ from a collection $D = \{d_1, ...d_N\}$.   

\subsubsection{Experimental Setup}

\paragraph{Metrics and Data} We benchmark the performance of popular retrieval methods and models in Table \ref{tab:ret-main} and evaluate with Recall@10, 100, 1000, and nDCG@10. We evaluate with 2851 \texttt{indirect single-removed} queries and provide ablation studies on the effect of query types and lengths in Table \ref{tab:retrieval-abla} and Figure \ref{fig:abla-lengh}. 

\paragraph{Late Interaction Models}
We run ColBERTv2 \citep{santhanam2022colbertv2} and long-context Jina-ColBERT \citep{gunther-etal-2023-jina} experiments with the PLAID-X implementation \citep{ecir2024translate-distill}. Due to the large size of the collection and GPU memory required by ColBERT indexing, we split \clercd\ and \clercp\ into four shards for processing. For ColBERTv2, which has a context limit of 512 tokens,\footnote{ColBERT generally assumes a max query token limit of 32, which we remove to incorporate our longer queries.} we conduct passage-level retrieval and aggregate the scores by taking the highest-ranking passage of a document to represent the overall document score (MaxP) \citep{Zhang2021ComparingSA}. For long-context ColBERT, we perform and evaluate document-level retrieval and truncate documents over the context lengths (2048, 4096).

\paragraph{Bi-Encoders}
We also benchmark the performance of state-of-the-art Bi-Encoders, including BGE-large-en-v1.5 \citep{bge_embedding}, E5-large-v2 \citep{Wang2022TextEB}, Contriever-MSMarco \citep{izacard2021contriever}. For DPR  models, we adopt the Tevatron pipeline \citep{Gao2022TevatronAE} and evaluate the base DPR model \citep{Karpukhin2020DensePR} trained on QA datasets. We also fine-tune BERT-base-uncased\footnote{\url{https://huggingface.co/jhu-clsp/BERT-DPR-CLERC-ft}} \citep{devlin2018bert} and LegalBERT\footnote{\url{https://huggingface.co/jhu-clsp/LegalBERT-DPR-CLERC-ft}} \citep{Holzenberger2020ADF} on 327K triples from \clercp\ for 10 epochs with a temperature of 0.01 and a learning rate of $10^{-4}$, with a contrastive learning-to-rank objective. 

\paragraph{Compute}
Please see Appendix \ref{app:compute} for the compute setup and hours of the experiments.

\subsubsection{Ablation Studies}
\label{exp:ret-abla}
\paragraph{Query Types}
We run retrieval experiments with BM25 on all four types of \clercs queries. Table \ref{tab:retrieval-abla} shows that \texttt{single-removed} are the easiest queries to retrieve, whereas \texttt{all-removed} queries remove non-central case citations and lose additional contextual information contained by them.

\paragraph{Query Length}
We also evaluate the effect of query length on BM25 across the range of 100 to 1000 words. Figure \ref{fig:abla-lengh} shows that longer query length improves recall and maximizes it at 300 words. Queries with more than 300 words are harder to retrieve, since as length increases, there are also more distracting contextual information about non-central citations that cause the retriever to find irrelevant passages (see Section \ref{sec:discuss-ret} for examples and qualitative analysis). Another limitation of increased query length is that it will raise the computational cost and storage, which is especially challenging to token-level late-interaction models like ColBERT \citep{khattab2020colbert}.
\subsubsection{Main Results and Analysis}
We show main results of the retrieval task in Table~\ref{tab:ret-main}. Overall, we see that \textbf{no zero-shot model scores higher than 50\% R@1K, indicating poor performance at retrieving case citations}. Due to the large size of the corpus, we find that late interaction models like ColBERTv2 \citep{santhanam2022colbertv2} perform the worst, followed by Bi-Encoder models which score around 42\% R@1K. BM25 performs the best at 48.3\% R@1K among all zero-shot models. We show later in Section~\ref{sec:discuss-ret} that this poor performance is due to domain shift to legal text and the struggles of models with long queries. 

\textbf{Domain shift can be alleviated through training}: fine-tuning BERT-base with the DPR architecture on \dataset's training set, we find a 36+ percent gain in R@1K, indicating that training on longer queries and legal text effectively improves performance in legal case retrieval, and it achieves the second highest performance after fine-tuned LegalBERT DPR, which is based on a model pretrained with legal texts \citep{Holzenberger2020ADF} and experiences less domain shift.

We also experiment with multiple Cross-Encoders and attach the results and analysis in \ref{app:cross-encoders}.

\subsection{Generating Legal Analyses from Citations}
\label{exp:arg-gen}
\paragraph{Task Formulation}
 Suppose a case document consisting of paragraphs ${\{p_1, p_2, \dots, p_N\}}$,
given ${R_t = \{r_1, r_2, ... r_T\}}$, a set of relevant passages or documents, and a prefix of case document $(p_1, p_2, \dots, p_{t-1})$,
a model is asked to continue to generate the next paragraph $\hat{p}_t$.
The generation quality is measured by the similarity between $p_t$ and $\hat{p}_t$.
In practice, the gold paragraph $p_t$ is randomly sampled from the last third of the paragraphs, i.e. $\floor{\frac{2N}{3}}\le t \le N-2$, and $R_t$ is selected as the citations used in $p_{t}$.

\subsubsection{Experimental Setup}
\paragraph{Citation Metrics}
To measure the faithfulness and degree of hallucination of legal analysis generation, we adopt Citation Recall (CR) and Citation Precision (CP) from \citet{Liu2023EvaluatingVI} and also propose Citation False Positive rate (CFP). Suppose a piece of generated text contains citations $\{c_1,\dots,c_M\}$ and a set of citations that should be cited, $C_r$, we define the following metrics:
\begin{align*}
\mathit{CR} &= \frac{\sum_i^{M} \mathbbm{1} \left[c_i \in C_r\right]}{|C_r|} \\
\mathit{CP} &= \frac{\sum_i^{M} \mathbbm{1} \left[c_i \in C_r\right]}{M} \\
\mathit{CFP} &= 1-\frac{\sum_i^{M} \mathbbm{1} \left[c_i \in C_r \vee \left(\bigvee_{j=1}^{t-1} c_i \sqsubseteq p_j\right) \right]}{M}
\end{align*}
where $\mathbbm{1}$ is the indicator function, and $a \sqsubseteq b$ indicates that $a$ is a substring of $b$. CR measures the fraction of relevant citations being generated, and CP measures the fraction of generated citations being relevant. CFP measures the fraction of the generated citations being hallucinated, which means that they come from unlisted sources. See an example calculation in Figure \ref{fig:eghalu}.

\paragraph{Metrics, Data, and Models}
We evaluate legal analysis generation with ROUGE \citep{Lin2004ROUGEAP}, BARTScore \citep{Yuan2021BARTScoreEG}, CP, CR, and CFP.

We run greedy generations on the test set (1k instances) of \clercg\ with Meta-Llama-3-8B-Instruct,\footnote{\url{https://github.com/meta-llama/llama3}} Mistral-7B-Instruct-v0.3 \citep{Jiang2023Mistral7}, GPT-4o \citep{gpt424}, and Gemma-1.1-7b-it \citep{gemini1.524}. 
We use FastChat~\citep{zheng2023judging} to host the open source models and the OpenAI API to access GPT-4o.\footnote{\url{https://platform.openai.com/playground}}
All the experiments are done using greedy decoding with a temperature of 1.
Readers may refer to Appendix \ref{app:arg-gen-prompt} for the prompts used.

\paragraph{Ablation study}
To study the effect of retrieved cases on generation, we perform an ablation study by not providing any citation documents to the model. 
The model is asked to generate the text solely based on the previous text, i.e. $(p_1, p_2, \dots, p_{t-1})$.

\subsubsection{Results and Analysis}
\label{sec:gen-res}
Table \ref{tab:arg-gen-res} shows results of analysis generation on \clercg. GPT-4o overall demonstrates the highest performance in terms of F-scores of ROUGE, CP, and CR, while Meta-Llama-3-8B-Instruct scores highest in BARTScore and second-highest across multiple metrics. 

\textbf{Under the full prompt, GPT-4o hallucinates the most}, as reflected by its CFP score (see Figure \ref{fig:eghalu}). Gemma-1.1-7b-it achieves both the minimum CFP with the full prompt and the worst CFP with the shortened prompt, which shows that it is most sensitive to prompting under our setting.

\textbf{Prompting with the texts of cited cases effectively boosts performance}, helping models cite more comprehensively, precisely, and with less hallucination, which is crucial for legal writing.

However, even with the full prompt, \textbf{generations have low CP and CR and considerably high CFP, which is concerning for legal applications}. Moreover, \textbf{CP, CR, and CFP overestimate analysis quality}. As we will discuss in Section \ref{sec:gen-metric}, high scores on the citation metrics do not imply that outputs are factual and analytically sound.

\section{Discussion}

\subsection{Limitations in Case Retrieval}
\label{sec:discuss-ret}
We analyze why popular retrieval models struggle on the \clercs retrieval task, as shown in Table \ref{tab:ret-main}. As an example, consider this shortened query:   
\begin{quote}
    \footnotesize
$\cdots$ (Omitted for brevity) $\cdots$
St. Paul Reins. Co., Ltd. v. Commercial Fin. Corp., 198 F.R.D. 508, 512 (N.D.Iowa 2000) (\underline{\textcolor{red4cb}{Boilerplate}} objections, including that a particul
ar document request was oppressive, burdensome and harassing, were “[i]n eve
ry respect ... textbook examples of what federal courts have routinely deeme
d to be improper objections.”) 
$\cdots$ (Omitted for brevity) $\cdots$
\underline{\textcolor{red4cb}{Elk Run}}’s Second Set of Interrogatories The court notes that Plaintiffs answers to Elk Run’s second set of interrogatories are utterly deficient because they are not answered under oath, signed by the person making the answers as required by Rule 33(b)(1) of the Federal Rules of Civil Procedure and \textbf{REDACTED} In addition to the deficiency outlined above, the court finds that the answers to interrogatory numbers 1 and 2 are inadequate, unresponsive, and inconsistent with the requirements of Rule 33. 
$\cdots$ (Omitted for brevity) $\cdots$
\end{quote}
where REDACTED is a placeholder for the removed central citation sentence, and we highlight \underline{\textcolor{red4cb}{underlined red words}} as ``distractor'' words. 

After examining the full texts of top-ranked results when searching with BM25, we discover that \textbf{models can be easily be ``distracted'' and retrieve non-central citations, usually due to the many overlaps of distractor words}. In the highest ranked case, \textit{228 F.R.D. 536}, the word ``Elk Run'' appears 45 times, which is the name of a sued company in the referenced case. In the second-highest case, \textit{309 F.R.D. 476}, the word ``Boilerplate'' appears 67 times, whereas ``Boilerplate objection'' is a common type of objection raised in court \citep{BaickerMcKeeFederalCivilRules}. These distractor words are often common words and legal jargon, which can have many occurrences in irrelevant documents and make it harder for popular models to retrieve. Hence, this domain shift (and our results in Table~\ref{tab:ret-main}) show how important fine-tuning on legal data is for legal retrieval models.

\subsection{Limitations in Analysis Generation and Metrics}
\label{sec:gen-metric}
As discussed in Section \ref{sec:gen-res}, the current models struggle on the citation metrics. However, we show an example to illustrate that \textbf{scoring high on current citation metrics does not imply good generation quality.} We sample and show a fragment of gold paragraph from 99 F. Supp. 2d 1115 (C.D. Cal. 1999) in Figure \ref{fig:egdoc}:
\begin{quote}
    \vspace{-1mm}
    \footnotesize
    \textit{\color{red4cb}{Divers’ status as a corporate officer does not exempt him from liability.}} \color{blue4cb}{See Kayes v. Pacific Lumber Co., 51 F.3d 1449, 1459 (9th Cir.1995)}
    \vspace{-1mm}
\end{quote}
where the \textit{\color{red4cb}{red italicized}} texts describe the analytical claim, supported by the case citation in \text{\color{blue4cb}{blue}}. On the other hand, Llama-3-8B outputs:
\begin{quote}
\vspace{-1mm}
    \footnotesize
    \textit{\color{red4cb}{As previously stated, 29 U.S.C. § 1002(21)(A) defines a fiduciary as a person who exercises any authority or control respecting management or disposition of [a plan’s] assets.}} \color{blue4cb}{P51 F.3d 1449, 1459 (9th Cir.1995).}
    \vspace{-1mm}
\end{quote}

While the generation makes the correct citation (``P51 F.3d 1449, 1459 (9th Cir.1995)'') and scores high on the citation metrics, \textbf{it fails to generate an analytical claim that can be supported by the citation}. From the gold paragraph, the citation should support the claim ``corporate officer has liability.'' However, the generated claim only defines what a fiduciary is, which was already defined in the previous context, rather than conveying useful new information for reasoning. See an additional example of GPT-4o hallucination in Appendix \ref{app:addeg}.

Therefore, we welcome the development of automated metrics that focus on evaluating the \textit{factual recall} aspect of text generation. One such approach could be extending metrics such as FActScore \citep{min-etal-2023-factscore} to the legal domain. However, as extracting the legal analyses itself is non-trivial in the legal domain, we leave this to future work.

\section{Conclusion}
Case law data is a high-quality  resource for building intelligence systems to assist legal professionals. Through an interdisciplinary effort with legal professionals, we transform the digitized corpus of case law \citep{cap} into \clerc, a dataset designed for training and evaluating models on long-context case retrieval and analysis generation tasks. We hope our work opens the gate for more exciting discoveries in AI built with legal text data.

\section{Limitations}
Our work provides an initial step into developing systems that can aid legal professionals. However, as we've shown, current systems are not well-suited to helping legal professionals on these tasks at the current moment, even those that are fine-tuned.

We use the legal data provided by CAP, which gathered the data through OCR. As such, our data may contain small OCR errors, although our analysis has found it to be generally of high quality.

\section*{Ethical Considerations}
In this work, we curate our dataset based on the corpus of historical U.S. federal case law. However, past case law might contain racisim, sexism, and all other ideologies that people have considered controversial or undesirable for society and people's well-being. Building legal intelligence systems on case law data has the risk of propagating such historical biases and ideologies, which we caution the NLP community to be aware of and encourage future endeavors to address.

\section*{Acknowledgement}

This work was supported in part by the U.S. National Science Foundation under grant 2204926. Opinions, findings, and conclusions or recommendations expressed in this article come from the authors and do not reflect the views of the National Science Foundation. We  alsothank Amy Wang for her advice on the visual designs in this paper.
\bibliography{custom,anthology}

\begin{thebibliography}{75}
\providecommand{\natexlab}[1]{#1}

\bibitem[{{Anthropic}(2024)}]{claude324}
{Anthropic}. 2024.
\newblock \href {https://www-cdn.anthropic.com/de8ba9b01c9ab7cbabf5c33b80b7bbc618857627/Model_Card_Claude_3.pdf} {The {{Claude}} 3 {{Model Family}}: {{Opus}}, {{Sonnet}}, {{Haiku}}}.

\bibitem[{Askari et~al.(2021)Askari, Verberne, Alonso, Marchesin, Najork, and Silvello}]{askari2021combining}
Arian Askari, Suzan Verberne, O~Alonso, S~Marchesin, M~Najork, and G~Silvello. 2021.
\newblock Combining lexical and neural retrieval with longformer-based summarization for effective case law retrieval.
\newblock In \emph{DESIRES}, pages 162--170.

\bibitem[{Baicker-McKee and Janssen(2022)}]{BaickerMcKeeFederalCivilRules}
Steven Baicker-McKee and William~M. Janssen. 2022.
\newblock \emph{Federal Civil Rules Handbook}.
\newblock Thomson Reuters.

\bibitem[{Beltagy et~al.(2020)Beltagy, Peters, and Cohan}]{longformer20}
Iz~Beltagy, Matthew~E. Peters, and Arman Cohan. 2020.
\newblock \href {https://arxiv.org/abs/2004.05150} {Longformer: {{The Long-Document Transformer}}}.

\bibitem[{Blair and Maron(1985)}]{blair1985evaluation}
David~C Blair and Melvin~E Maron. 1985.
\newblock An evaluation of retrieval effectiveness for a full-text document-retrieval system.
\newblock \emph{Communications of the ACM}, 28(3):289--299.

\bibitem[{Brown et~al.(2020)Brown, Mann, Ryder, Subbiah, Kaplan, Dhariwal, Neelakantan, Shyam, Sastry, Askell, Agarwal, Herbert-Voss, Krueger, Henighan, Child, Ramesh, Ziegler, Wu, Winter, Hesse, Chen, Sigler, Litwin, Gray, Chess, Clark, Berner, McCandlish, Radford, Sutskever, and Amodei}]{brown2020language}
Tom~B. Brown, Benjamin Mann, Nick Ryder, Melanie Subbiah, Jared Kaplan, Prafulla Dhariwal, Arvind Neelakantan, Pranav Shyam, Girish Sastry, Amanda Askell, Sandhini Agarwal, Ariel Herbert-Voss, Gretchen Krueger, Tom Henighan, Rewon Child, Aditya Ramesh, Daniel~M. Ziegler, Jeffrey Wu, Clemens Winter, Christopher Hesse, Mark Chen, Eric Sigler, Mateusz Litwin, Scott Gray, Benjamin Chess, Jack Clark, Christopher Berner, Sam McCandlish, Alec Radford, Ilya Sutskever, and Dario Amodei. 2020.
\newblock \href {https://arxiv.org/abs/2005.14165} {Language models are few-shot learners}.

\bibitem[{Bulatov et~al.(2022)Bulatov, Kuratov, and Burtsev}]{rmt22}
Aydar Bulatov, Yuri Kuratov, and Mikhail~S. Burtsev. 2022.
\newblock \href {https://arxiv.org/abs/2207.06881} {Recurrent {{Memory Transformer}}}.
\newblock In \emph{Proceedings of Conference on Neural Information Processing Systems (NeurIPS)}.

\bibitem[{Cao et~al.(2024)Cao, Wang, Xiao, and Sun}]{cao2024PILOTLegalCase}
Lang Cao, Zifeng Wang, Cao Xiao, and Jimeng Sun. 2024.
\newblock \href {https://arxiv.org/abs/2401.15770} {{{PILOT}}: {{Legal Case Outcome Prediction}} with {{Case Law}}}.

\bibitem[{CAP(2024)}]{cap}
CAP. 2024.
\newblock \href {https://case.law} {Caselaw access project}.

\bibitem[{Chen et~al.(2023)Chen, Wong, Chen, and Tian}]{chen2023ExtendingContextWindow}
Shouyuan Chen, Sherman Wong, Liangjian Chen, and Yuandong Tian. 2023.
\newblock \href {https://arxiv.org/abs/2306.15595} {Extending {{Context Window}} of {{Large Language Models}} via {{Positional Interpolation}}}.

\bibitem[{Choromanski et~al.(2021)Choromanski, Likhosherstov, Dohan, Song, Gane, Sarlos, Hawkins, Davis, Mohiuddin, Kaiser, Belanger, Colwell, and Weller}]{performer21}
Krzysztof Choromanski, Valerii Likhosherstov, David Dohan, Xingyou Song, Andreea Gane, Tamas Sarlos, Peter Hawkins, Jared Davis, Afroz Mohiuddin, Lukasz Kaiser, David Belanger, Lucy Colwell, and Adrian Weller. 2021.
\newblock \href {https://arxiv.org/abs/2009.14794} {Rethinking {{Attention}} with {{Performers}}}.
\newblock In \emph{Proceedings of International Conference on Learning Representations (ICLR)}.

\bibitem[{Cushman et~al.(2021)Cushman, Dahl, and Lissner}]{eyecite}
Jack Cushman, Matthew Dahl, and Michael Lissner. 2021.
\newblock \href {https://doi.org/10.21105/joss.03617} {eyecite: A tool for parsing legal citations}.
\newblock \emph{Journal of Open Source Software}, 6(66):3617.

\bibitem[{Dahl et~al.(2024)Dahl, Magesh, Suzgun, and Ho}]{Dahl2024LargeLF}
Matthew Dahl, Varun Magesh, Mirac Suzgun, and Daniel~E. Ho. 2024.
\newblock \href {https://api.semanticscholar.org/CorpusID:266725450} {Large legal fictions: Profiling legal hallucinations in large language models}.
\newblock \emph{ArXiv}, abs/2401.01301.

\bibitem[{Dai et~al.(2019)Dai, Yang, Yang, Carbonell, Le, and Salakhutdinov}]{dai-etal-2019-transformer}
Zihang Dai, Zhilin Yang, Yiming Yang, Jaime Carbonell, Quoc Le, and Ruslan Salakhutdinov. 2019.
\newblock \href {https://doi.org/10.18653/v1/P19-1285} {Transformer-{XL}: Attentive language models beyond a fixed-length context}.
\newblock In \emph{Proceedings of the 57th Annual Meeting of the Association for Computational Linguistics}, pages 2978--2988, Florence, Italy. Association for Computational Linguistics.

\bibitem[{Devlin et~al.(2018)Devlin, Chang, Lee, and Toutanova}]{devlin2018bert}
Jacob Devlin, Ming-Wei Chang, Kenton Lee, and Kristina Toutanova. 2018.
\newblock Bert: Pre-training of deep bidirectional transformers for language understanding.
\newblock \emph{arXiv preprint arXiv:1810.04805}.

\bibitem[{Dong et~al.(2024)Dong, Tang, Li, Zhao, and Wen}]{dong2024BAMBOOComprehensiveBenchmark}
Zican Dong, Tianyi Tang, Junyi Li, Wayne~Xin Zhao, and Ji-Rong Wen. 2024.
\newblock \href {https://arxiv.org/abs/2309.13345} {{{BAMBOO}}: {{A Comprehensive Benchmark}} for {{Evaluating Long Text Modeling Capacities}} of {{Large Language Models}}}.
\newblock In \emph{Proceedings of Language Resources and Evaluation Conference (LREC)}.

\bibitem[{Ebietomere and Ekuobase(2019)}]{ebietomere2019semantic}
Esingbemi~Princewill Ebietomere and Godspower~Osaretin Ekuobase. 2019.
\newblock A semantic retrieval system for case law.
\newblock \emph{Applied Computer Systems}, 24(1):38--48.

\bibitem[{Fu et~al.(2023)Fu, Arora, Grogan, Johnson, Eyuboglu, Thomas, Spector, Poli, Rudra, and R{\'e}}]{fu2023monarch}
Daniel~Y Fu, Simran Arora, Jessica Grogan, Isys Johnson, Sabri Eyuboglu, Armin~W Thomas, Benjamin Spector, Michael Poli, Atri Rudra, and Christopher R{\'e}. 2023.
\newblock Monarch mixer: A simple sub-quadratic gemm-based architecture.
\newblock In \emph{Advances in Neural Information Processing Systems}.

\bibitem[{Gao et~al.(2022)Gao, Ma, Lin, and Callan}]{Gao2022TevatronAE}
Luyu Gao, Xueguang Ma, Jimmy~J. Lin, and Jamie Callan. 2022.
\newblock Tevatron: An efficient and flexible toolkit for dense retrieval.
\newblock \emph{ArXiv}, abs/2203.05765.

\bibitem[{Gesnouin et~al.(2024)Gesnouin, Tannier, Silva, Tapory, Brier, Simon, Rozenberg, Woehrel, Yakaabi, Binder, Marie, Caron, Nogueira, Fontas, Puydebois, Theophile, Morandi, Petit, Creissac, Ennouchy, Valetoux, Visade, Balloux, Cortes, Devineau, Tan, Namara, and Yang}]{Gesnouin2024LLaMandementLL}
Joseph Gesnouin, Yannis Tannier, Christophe Gomes~Da Silva, Hatim Tapory, Camille Brier, Hugo Simon, Raphael Rozenberg, Hermann Woehrel, Mehdi~El Yakaabi, Thomas Binder, Guillaume Marie, Emilie Caron, Mathile Nogueira, Thomas Fontas, Laure Puydebois, Marie Theophile, Stephane Morandi, Mael Petit, David Creissac, Pauline Ennouchy, Elise Valetoux, Celine Visade, Severine Balloux, Emmanuel Cortes, Pierre-Etienne Devineau, Ulrich Tan, Esther~Mac Namara, and Su~Yang. 2024.
\newblock \href {https://api.semanticscholar.org/CorpusID:267311712} {Llamandement: Large language models for summarization of french legislative proposals}.
\newblock \emph{ArXiv}, abs/2401.16182.

\bibitem[{Goebel et~al.(2024)Goebel, Kano, Kim, Rabelo, Satoh, and Yoshioka}]{goebel2023coliee}
Randy Goebel, Yoshinobu Kano, Mi-Young Kim, Juliano Rabelo, Ken Satoh, and Masaharu Yoshioka. 2024.
\newblock Overview and discussion of the competition on legal information, extraction/entailment (coliee) 2023.
\newblock \emph{The Review of Socionetwork Strategies}.

\bibitem[{Guha et~al.(2023)Guha, Nyarko, Ho, Ré, Chilton, Narayana, Chohlas-Wood, Peters, Waldon, Rockmore, Zambrano, Talisman, Hoque, Surani, Fagan, Sarfaty, Dickinson, Porat, Hegland, Wu, Nudell, Niklaus, Nay, Choi, Tobia, Hagan, Ma, Livermore, Rasumov-Rahe, Holzenberger, Kolt, Henderson, Rehaag, Goel, Gao, Williams, Gandhi, Zur, Iyer, and Li}]{guha2023legalbench}
Neel Guha, Julian Nyarko, Daniel~E. Ho, Christopher Ré, Adam Chilton, Aditya Narayana, Alex Chohlas-Wood, Austin Peters, Brandon Waldon, Daniel~N. Rockmore, Diego Zambrano, Dmitry Talisman, Enam Hoque, Faiz Surani, Frank Fagan, Galit Sarfaty, Gregory~M. Dickinson, Haggai Porat, Jason Hegland, Jessica Wu, Joe Nudell, Joel Niklaus, John Nay, Jonathan~H. Choi, Kevin Tobia, Margaret Hagan, Megan Ma, Michael Livermore, Nikon Rasumov-Rahe, Nils Holzenberger, Noam Kolt, Peter Henderson, Sean Rehaag, Sharad Goel, Shang Gao, Spencer Williams, Sunny Gandhi, Tom Zur, Varun Iyer, and Zehua Li. 2023.
\newblock \href {https://arxiv.org/abs/2308.11462} {Legalbench: A collaboratively built benchmark for measuring legal reasoning in large language models}.
\newblock \emph{Preprint}, arXiv:2308.11462.

\bibitem[{G{\"u}nther et~al.(2023)G{\"u}nther, Mastrapas, Wang, Xiao, and Geuter}]{gunther-etal-2023-jina}
Michael G{\"u}nther, Georgios Mastrapas, Bo~Wang, Han Xiao, and Jonathan Geuter. 2023.
\newblock \href {https://doi.org/10.18653/v1/2023.nlposs-1.2} {{J}ina embeddings: A novel set of high-performance sentence embedding models}.
\newblock In \emph{Proceedings of the 3rd Workshop for Natural Language Processing Open Source Software (NLP-OSS 2023)}, pages 8--18, Singapore. Association for Computational Linguistics.

\bibitem[{Henderson et~al.(2022)Henderson, Krass, Zheng, Guha, Manning, Jurafsky, and Ho}]{hendersonkrass2022pileoflaw}
Peter Henderson, Mark Krass, Lucia Zheng, Neel Guha, Christopher Manning, Dan Jurafsky, and Daniel~E Ho. 2022.
\newblock Pile of law: Learning responsible data filtering from the law and a 256gb open-source legal dataset.

\bibitem[{Holzenberger et~al.(2020)Holzenberger, Blair-Stanek, and Durme}]{Holzenberger2020ADF}
Nils Holzenberger, Andrew Blair-Stanek, and Benjamin~Van Durme. 2020.
\newblock \href {https://api.semanticscholar.org/CorpusID:218581117} {A dataset for statutory reasoning in tax law entailment and question answering}.
\newblock In \emph{NLLP@KDD}.

\bibitem[{Izacard et~al.(2021)Izacard, Caron, Hosseini, Riedel, Bojanowski, Joulin, and Grave}]{izacard2021contriever}
Gautier Izacard, Mathilde Caron, Lucas Hosseini, Sebastian Riedel, Piotr Bojanowski, Armand Joulin, and Edouard Grave. 2021.
\newblock \href {https://doi.org/10.48550/ARXIV.2112.09118} {Unsupervised dense information retrieval with contrastive learning}.

\bibitem[{Jiang et~al.(2023)Jiang, Sablayrolles, Mensch, Bamford, Chaplot, de~Las~Casas, Bressand, Lengyel, Lample, Saulnier, Lavaud, Lachaux, Stock, Scao, Lavril, Wang, Lacroix, and Sayed}]{Jiang2023Mistral7}
Albert~Qiaochu Jiang, Alexandre Sablayrolles, Arthur Mensch, Chris Bamford, Devendra~Singh Chaplot, Diego de~Las~Casas, Florian Bressand, Gianna Lengyel, Guillaume Lample, Lucile Saulnier, L'elio~Renard Lavaud, Marie-Anne Lachaux, Pierre Stock, Teven~Le Scao, Thibaut Lavril, Thomas Wang, Timoth{\'e}e Lacroix, and William~El Sayed. 2023.
\newblock \href {https://api.semanticscholar.org/CorpusID:263830494} {Mistral 7b}.
\newblock \emph{ArXiv}, abs/2310.06825.

\bibitem[{Jiang et~al.(2024)Jiang, Zhang, Mahari, Kessler, Ma, August, Li, Pentland, Kim, Kabbara, and Roy}]{Jiang2024LeveragingLL}
Hang Jiang, Xiajie Zhang, Robert Mahari, Daniel Kessler, Eric Ma, Tal August, Irene Li, Alex~'Sandy' Pentland, Yoon Kim, Jad Kabbara, and Deb Roy. 2024.
\newblock \href {https://api.semanticscholar.org/CorpusID:268031822} {Leveraging large language models for learning complex legal concepts through storytelling}.
\newblock \emph{ArXiv}, abs/2402.17019.

\bibitem[{{Josh Achiam} et~al.(2024){Josh Achiam}, {Steven Adler}, {Sandhini Agarwal}, and {et al.}}]{gpt424}
{Josh Achiam}, {Steven Adler}, {Sandhini Agarwal}, and {et al.} 2024.
\newblock \href {https://arxiv.org/abs/2303.08774} {{{GPT-4 Technical Report}}}.

\bibitem[{Karpukhin et~al.(2020{\natexlab{a}})Karpukhin, Oguz, Min, Lewis, Wu, Edunov, Chen, and Yih}]{karpukhin-etal-2020-dense}
Vladimir Karpukhin, Barlas Oguz, Sewon Min, Patrick Lewis, Ledell Wu, Sergey Edunov, Danqi Chen, and Wen-tau Yih. 2020{\natexlab{a}}.
\newblock \href {https://doi.org/10.18653/v1/2020.emnlp-main.550} {Dense passage retrieval for open-domain question answering}.
\newblock In \emph{Proceedings of the 2020 Conference on Empirical Methods in Natural Language Processing (EMNLP)}, pages 6769--6781, Online. Association for Computational Linguistics.

\bibitem[{Karpukhin et~al.(2020{\natexlab{b}})Karpukhin, Oğuz, Min, Lewis, Wu, Edunov, Chen, and tau Yih}]{Karpukhin2020DensePR}
Vladimir Karpukhin, Barlas Oğuz, Sewon Min, Patrick Lewis, Ledell~Yu Wu, Sergey Edunov, Danqi Chen, and Wen tau Yih. 2020{\natexlab{b}}.
\newblock \href {https://api.semanticscholar.org/CorpusID:215737187} {Dense passage retrieval for open-domain question answering}.
\newblock \emph{ArXiv}, abs/2004.04906.

\bibitem[{Khattab and Zaharia(2020)}]{khattab2020colbert}
Omar Khattab and Matei Zaharia. 2020.
\newblock \href {https://doi.org/10.1145/3397271.3401075} {Colbert: Efficient and effective passage search via contextualized late interaction over {BERT}}.
\newblock In \emph{Proceedings of the 43rd International {ACM} {SIGIR} conference on research and development in Information Retrieval, {SIGIR} 2020, Virtual Event, China, July 25-30, 2020}, pages 39--48. {ACM}.

\bibitem[{Lewis et~al.(2020)Lewis, Perez, Piktus, Petroni, Karpukhin, Goyal, Kuttler, Lewis, Yih, Rocktaschel, Riedel, and Kiela}]{rag20}
Patrick Lewis, Ethan Perez, Aleksandra Piktus, Fabio Petroni, Vladimir Karpukhin, Naman Goyal, Heinrich Kuttler, Mike Lewis, Wen-tau Yih, Tim Rocktaschel, Sebastian Riedel, and Douwe Kiela. 2020.
\newblock \href {https://arxiv.org/abs/2005.11401} {Retrieval-{{Augmented Generation}} for {{Knowledge-Intensive NLP Tasks}}}.
\newblock In \emph{Proceedings of Conference on Neural Information Processing Systems (NeurIPS)}.

\bibitem[{Li et~al.(2023)Li, Shao, Xie, Sheng, Zheng, Gonzalez, Stoica, Ma, and Zhang}]{longchat23}
Dacheng Li, Rulin Shao, Anze Xie, Ying Sheng, Lianmin Zheng, Joseph~E Gonzalez, Ion Stoica, Xuezhe Ma, and Hao Zhang. 2023.
\newblock \href {https://openreview.net/pdf?id=LywifFNXV5} {How {{Long Can Context Length}} of {{Open-Source LLMs}} truly {{Promise}}?}
\newblock In \emph{Proceedings of Workshop on Instruction Tuning and Instruction Following}.

\bibitem[{Li et~al.(2024)Li, Shao, Wu, Ai, Ma, and Liu}]{li2023lecardv2}
Haitao Li, Yunqiu Shao, Yueyue Wu, Qingyao Ai, Yixiao Ma, and Yiqun Liu. 2024.
\newblock \href {https://arxiv.org/abs/2310.17609} {Lecardv2: A large-scale chinese legal case retrieval dataset}.
\newblock \emph{ArXiv preprint}.

\bibitem[{Li and Zhang(2021)}]{Li2021CourtOG}
Quanzhi Li and Qiong Zhang. 2021.
\newblock \href {https://api.semanticscholar.org/CorpusID:235363775} {Court opinion generation from case fact description with legal basis}.
\newblock In \emph{AAAI Conference on Artificial Intelligence}.

\bibitem[{Lin(2004)}]{Lin2004ROUGEAP}
Chin-Yew Lin. 2004.
\newblock \href {https://api.semanticscholar.org/CorpusID:964287} {Rouge: A package for automatic evaluation of summaries}.
\newblock In \emph{Annual Meeting of the Association for Computational Linguistics}.

\bibitem[{Lin and Cheng(2024)}]{Lin2024LegalDD}
Chun-Hsien Lin and Pu-Jen Cheng. 2024.
\newblock \href {https://api.semanticscholar.org/CorpusID:269505472} {Legal documents drafting with fine-tuned pre-trained large language model}.
\newblock \emph{Software Engineering \& Trends}.

\bibitem[{Lippmann(2018)}]{lippmann_criminal_procedure}
Matthew Lippmann. 2018.
\newblock \emph{Criminal Procedure}, 3rd edition.
\newblock SAGE Publications, Inc, Thousand Oaks, CA.

\bibitem[{Liu et~al.(2024)Liu, Lin, Hewitt, Paranjape, Bevilacqua, Petroni, and Liang}]{lost24}
Nelson~F. Liu, Kevin Lin, John Hewitt, Ashwin Paranjape, Michele Bevilacqua, Fabio Petroni, and Percy Liang. 2024.
\newblock \href {https://doi.org/10.1162/tacl_a_00638} {Lost in the {{Middle}}: {{How Language Models Use Long Contexts}}}.
\newblock \emph{Transactions of the Association for Computational Linguistics (TACL)}, pages 157--173.

\bibitem[{Liu et~al.(2023)Liu, Zhang, and Liang}]{Liu2023EvaluatingVI}
Nelson~F. Liu, Tianyi Zhang, and Percy Liang. 2023.
\newblock \href {https://api.semanticscholar.org/CorpusID:258212854} {Evaluating verifiability in generative search engines}.
\newblock \emph{ArXiv}, abs/2304.09848.

\bibitem[{Locke and Zuccon(2022)}]{locke2022case}
Daniel Locke and Guido Zuccon. 2022.
\newblock Case law retrieval: problems, methods, challenges and evaluations in the last 20 years.
\newblock \emph{arXiv preprint arXiv:2202.07209}.

\bibitem[{Ma et~al.(2023)Ma, Wang, Yang, Wei, and Lin}]{Ma2023FineTuningLF}
Xueguang Ma, Liang Wang, Nan Yang, Furu Wei, and Jimmy Lin. 2023.
\newblock \href {https://api.semanticscholar.org/CorpusID:263908865} {Fine-tuning llama for multi-stage text retrieval}.
\newblock \emph{ArXiv}, abs/2310.08319.

\bibitem[{Ma et~al.(2021)Ma, Shao, Wu, Liu, Zhang, Zhang, and Ma}]{ma2021lecard}
Yixiao Ma, Yunqiu Shao, Yueyue Wu, Yiqun Liu, Ruizhe Zhang, Min Zhang, and Shaoping Ma. 2021.
\newblock Lecard: A legal case retrieval dataset for chinese law system.
\newblock In \emph{Proceedings of the 44th International ACM SIGIR Conference on Research and Development in Information Retrieval}, pages 2342--2348.

\bibitem[{Magesh et~al.(2024)Magesh, Surani, Dahl, Suzgun, Manning, and Ho}]{Magesh2024HallucinationFreeAT}
Varun Magesh, Faiz Surani, Matthew Dahl, Mirac Suzgun, Christopher~D. Manning, and Daniel~E. Ho. 2024.
\newblock \href {https://api.semanticscholar.org/CorpusID:269976547} {Hallucination-free? assessing the reliability of leading ai legal research tools}.

\bibitem[{Mahari(2021)}]{Mahari2021AutoLAWAL}
Robert Mahari. 2021.
\newblock \href {https://api.semanticscholar.org/CorpusID:235683267} {Autolaw: Augmented legal reasoning through legal precedent prediction}.
\newblock \emph{ArXiv}, abs/2106.16034.

\bibitem[{Mahari et~al.(2023)Mahari, Stammbach, Ash, and Pentland}]{mahari2023LePaRD}
Robert Mahari, Dominik Stammbach, Elliott Ash, and Alex'Sandy' Pentland. 2023.
\newblock Lepard: A large-scale dataset of judges citing precedents.
\newblock \emph{arXiv preprint}.

\bibitem[{Min et~al.(2023)Min, Krishna, Lyu, Lewis, Yih, Koh, Iyyer, Zettlemoyer, and Hajishirzi}]{min-etal-2023-factscore}
Sewon Min, Kalpesh Krishna, Xinxi Lyu, Mike Lewis, Wen-tau Yih, Pang Koh, Mohit Iyyer, Luke Zettlemoyer, and Hannaneh Hajishirzi. 2023.
\newblock \href {https://doi.org/10.18653/v1/2023.emnlp-main.741} {{FA}ct{S}core: Fine-grained atomic evaluation of factual precision in long form text generation}.
\newblock In \emph{Proceedings of the 2023 Conference on Empirical Methods in Natural Language Processing}, pages 12076--12100, Singapore. Association for Computational Linguistics.

\bibitem[{Nogueira et~al.(2020)Nogueira, Jiang, Pradeep, and Lin}]{Nogueira2020DocumentRW}
Rodrigo Nogueira, Zhiying Jiang, Ronak Pradeep, and Jimmy~J. Lin. 2020.
\newblock \href {https://api.semanticscholar.org/CorpusID:212725651} {Document ranking with a pretrained sequence-to-sequence model}.
\newblock In \emph{Findings}.

\bibitem[{Qin et~al.(2023)Qin, Feng, and Van~Durme}]{qin-etal-2023-nlp}
Guanghui Qin, Yukun Feng, and Benjamin Van~Durme. 2023.
\newblock \href {https://doi.org/10.18653/v1/2023.eacl-main.273} {The {NLP} task effectiveness of long-range transformers}.
\newblock In \emph{Proceedings of the 17th Conference of the European Chapter of the Association for Computational Linguistics}, pages 3774--3790, Dubrovnik, Croatia. Association for Computational Linguistics.

\bibitem[{Qin et~al.(2024{\natexlab{a}})Qin, Rosset, Chau, Rao, and Van~Durme}]{qin-etal-dodo-2024}
Guanghui Qin, Corby Rosset, Ethan~C Chau, Nikhil Rao, and Benjamin Van~Durme. 2024{\natexlab{a}}.
\newblock \href {https://arxiv.org/pdf/2310.02409} {Dodo: Dynamic contextual compression for decoder-only lms}.
\newblock In \emph{Proceedings of the 2024 Conference on Association for Computational Linguistics (ACL)}.

\bibitem[{Qin et~al.(2024{\natexlab{b}})Qin, Cao, Yu, Si, Chen, and Xu}]{qin2024ExplicitlyIntegratingJudgment}
Weicong Qin, Zelin Cao, Weijie Yu, Zihua Si, Sirui Chen, and Jun Xu. 2024{\natexlab{b}}.
\newblock \href {https://doi.org/10.1145/3626772.3657717} {Explicitly {{Integrating Judgment Prediction}} with {{Legal Document Retrieval}}: {{A Law-Guided Generative Approach}}}.

\bibitem[{Reid et~al.(2024)Reid, Savinov, Teplyashin, and {et al.}}]{gemini1.524}
Machel Reid, Nikolay Savinov, Denis Teplyashin, and {et al.} 2024.
\newblock \href {https://arxiv.org/abs/2403.05530} {Gemini 1.5: {{Unlocking}} multimodal understanding across millions of tokens of context}.

\bibitem[{Ren et~al.(2021)Ren, Qu, Liu, Zhao, She, Wu, Wang, and Wen}]{ren2021rocketqav2}
Ruiyang Ren, Yingqi Qu, Jing Liu, Wayne~Xin Zhao, QiaoQiao She, Hua Wu, Haifeng Wang, and Ji-Rong Wen. 2021.
\newblock \href {https://doi.org/10.18653/v1/2021.emnlp-main.224} {{R}ocket{QA}v2: A joint training method for dense passage retrieval and passage re-ranking}.
\newblock In \emph{Proceedings of the 2021 Conference on Empirical Methods in Natural Language Processing}, pages 2825--2835, Online and Punta Cana, Dominican Republic. Association for Computational Linguistics.

\bibitem[{Review(2020)}]{bluebook}
Columbia~Law Review. 2020.
\newblock \emph{The Bluebook: A Uniform System of Citation}.
\newblock The Harvard LawReview Association Gannett House, 1511 Massachusetts Avenue Cambridge, Massachusetts 02138 U.S.A.

\bibitem[{Saad-Falcon et~al.(2024)Saad-Falcon, Fu, Arora, Guha, and Ré}]{saad-falcon2024long}
Jon Saad-Falcon, Daniel~Y. Fu, Simran Arora, Neel Guha, and Christopher Ré. 2024.
\newblock Benchmarking and building long-context retrieval models with loco and m2-bert.
\newblock \emph{arXiv preprint arXiv:2402.07440}.

\bibitem[{Santhanam et~al.(2022)Santhanam, Khattab, Saad-Falcon, Potts, and Zaharia}]{santhanam2022colbertv2}
Keshav Santhanam, Omar Khattab, Jon Saad-Falcon, Christopher Potts, and Matei Zaharia. 2022.
\newblock \href {https://doi.org/10.18653/v1/2022.naacl-main.272} {{C}ol{BERT}v2: Effective and efficient retrieval via lightweight late interaction}.
\newblock In \emph{Proceedings of the 2022 Conference of the North American Chapter of the Association for Computational Linguistics: Human Language Technologies}, pages 3715--3734, Seattle, United States. Association for Computational Linguistics.

\bibitem[{Santosh et~al.(2024)Santosh, Haddad, and Grabmair}]{santosh24ecthr}
T.~Y. S.~S. Santosh, Rashid Haddad, and Matthias Grabmair. 2024.
\newblock \href {https://aclanthology.org/2024.lrec-main.486} {Ecthr-pcr: {A} dataset for precedent understanding and prior case retrieval in the european court of human rights}.
\newblock In \emph{Proceedings of the 2024 Joint International Conference on Computational Linguistics, Language Resources and Evaluation, {LREC/COLING} 2024, 20-25 May, 2024, Torino, Italy}, pages 5473--5483. {ELRA} and {ICCL}.

\bibitem[{Shaham et~al.(2022)Shaham, Segal, Ivgi, Efrat, Yoran, Haviv, Gupta, Xiong, Geva, Berant, and Levy}]{shaham-etal-2022-scrolls}
Uri Shaham, Elad Segal, Maor Ivgi, Avia Efrat, Ori Yoran, Adi Haviv, Ankit Gupta, Wenhan Xiong, Mor Geva, Jonathan Berant, and Omer Levy. 2022.
\newblock \href {https://doi.org/10.18653/v1/2022.emnlp-main.823} {{SCROLLS}: Standardized {C}ompa{R}ison over long language sequences}.
\newblock In \emph{Proceedings of the 2022 Conference on Empirical Methods in Natural Language Processing}, pages 12007--12021, Abu Dhabi, United Arab Emirates. Association for Computational Linguistics.

\bibitem[{Shen et~al.(2022)Shen, Lo, Yu, Dahlberg, Schlanger, and Downey}]{Shen2022MultiLexSumRS}
Zejiang Shen, Kyle Lo, Lauren~Jane Yu, Nathan Dahlberg, Margo Schlanger, and Doug Downey. 2022.
\newblock \href {https://api.semanticscholar.org/CorpusID:249927023} {Multi-lexsum: Real-world summaries of civil rights lawsuits at multiple granularities}.
\newblock \emph{ArXiv}, abs/2206.10883.

\bibitem[{Su et~al.(2024)Su, Ahmed, Lu, Pan, Bo, and Liu}]{roformer24}
Jianlin Su, Murtadha Ahmed, Yu~Lu, Shengfeng Pan, Wen Bo, and Yunfeng Liu. 2024.
\newblock \href {https://doi.org/10.1016/j.neucom.2023.127063} {{{RoFormer}}: {{Enhanced}} transformer with {{Rotary Position Embedding}}}.
\newblock \emph{Neurocomputing}, page 127063.

\bibitem[{Tay et~al.(2020)Tay, Dehghani, Abnar, Shen, Bahri, Pham, Rao, Yang, Ruder, and Metzler}]{tay2020LongRangeArena}
Yi~Tay, Mostafa Dehghani, Samira Abnar, Yikang Shen, Dara Bahri, Philip Pham, Jinfeng Rao, Liu Yang, Sebastian Ruder, and Donald Metzler. 2020.
\newblock \href {https://arxiv.org/abs/2011.04006} {Long {{Range Arena}}: {{A Benchmark}} for {{Efficient Transformers}}}.

\bibitem[{Thakur et~al.(2021)Thakur, Reimers, R{\"u}ckl{\'e}, Srivastava, and Gurevych}]{thakur2021beir}
Nandan Thakur, Nils Reimers, Andreas R{\"u}ckl{\'e}, Abhishek Srivastava, and Iryna Gurevych. 2021.
\newblock \href {https://arxiv.org/abs/2104.08663} {Beir: A heterogenous benchmark for zero-shot evaluation of information retrieval models}.
\newblock \emph{ArXiv preprint}, abs/2104.08663.

\bibitem[{Touvron et~al.(2023)Touvron, Lavril, Izacard, Martinet, Lachaux, Lacroix, Rozi{\`e}re, Goyal, Hambro, Azhar, Rodriguez, Joulin, Grave, and Lample}]{touvron2023llama}
Hugo Touvron, Thibaut Lavril, Gautier Izacard, Xavier Martinet, Marie-Anne Lachaux, Timoth{\'e}e Lacroix, Baptiste Rozi{\`e}re, Naman Goyal, Eric Hambro, Faisal Azhar, Aurelien Rodriguez, Armand Joulin, Edouard Grave, and Guillaume Lample. 2023.
\newblock Llama: Open and efficient foundation language models.
\newblock \emph{arXiv preprint arXiv:2302.13971}.

\bibitem[{Tuvey and Sen(2023)}]{tuvey2023argument}
Oscar Tuvey and Procheta Sen. 2023.
\newblock Automated argument generation from legal facts.
\newblock \emph{arXiv:2310.05680}.

\bibitem[{T.Y.S.S et~al.(2024)T.Y.S.S, Haddad, and Grabmair}]{ecthr}
Santosh T.Y.S.S, Rashid~Gustav Haddad, and Matthias Grabmair. 2024.
\newblock \href {https://arxiv.org/abs/2404.00596} {Ecthr-pcr: A dataset for precedent understanding and prior case retrieval in the european court of human rights}.
\newblock \emph{ArXiv preprint}.

\bibitem[{Wang et~al.(2022)Wang, Yang, Huang, Jiao, Yang, Jiang, Majumder, and Wei}]{Wang2022TextEB}
Liang Wang, Nan Yang, Xiaolong Huang, Binxing Jiao, Linjun Yang, Daxin Jiang, Rangan Majumder, and Furu Wei. 2022.
\newblock \href {https://api.semanticscholar.org/CorpusID:254366618} {Text embeddings by weakly-supervised contrastive pre-training}.
\newblock \emph{ArXiv}, abs/2212.03533.

\bibitem[{Wang et~al.(2023)Wang, Yang, Huang, Yang, Majumder, and Wei}]{wang2023improving}
Liang Wang, Nan Yang, Xiaolong Huang, Linjun Yang, Rangan Majumder, and Furu Wei. 2023.
\newblock Improving text embeddings with large language models.
\newblock \emph{arXiv preprint arXiv:2401.00368}.

\bibitem[{Weller et~al.(2024)Weller, Chang, MacAvaney, Lo, Cohan, Durme, Lawrie, and Soldaini}]{Weller2024FollowIREA}
Orion Weller, Benjamin Chang, Sean MacAvaney, Kyle Lo, Arman Cohan, Benjamin~Van Durme, Dawn Lawrie, and Luca Soldaini. 2024.
\newblock \href {https://api.semanticscholar.org/CorpusID:268667440} {Followir: Evaluating and teaching information retrieval models to follow instructions}.
\newblock \emph{ArXiv}, abs/2403.15246.

\bibitem[{Wilson(1962)}]{wilson1962computer}
Robert~A Wilson. 1962.
\newblock Computer retrieval of case law.
\newblock \emph{Sw. LJ}, 16:409.

\bibitem[{Xiao et~al.(2023)Xiao, Liu, Zhang, and Muennighoff}]{bge_embedding}
Shitao Xiao, Zheng Liu, Peitian Zhang, and Niklas Muennighoff. 2023.
\newblock \href {https://arxiv.org/abs/2309.07597} {C-pack: Packaged resources to advance general chinese embedding}.
\newblock \emph{Preprint}, arXiv:2309.07597.

\bibitem[{Yang et~al.(2024)Yang, Lawrie, Mayfield, Oard, and Miller}]{ecir2024translate-distill}
Eugene Yang, Dawn Lawrie, James Mayfield, Douglas~W. Oard, and Scott Miller. 2024.
\newblock \href {https://arxiv.org/abs/2401.04810} {Translate-distill: Learning cross-language dense retrieval by translation and distillation}.
\newblock In \emph{Proceedings of the 46th European Conference on Information Retrieval (ECIR)}.

\bibitem[{Yuan et~al.(2021)Yuan, Neubig, and Liu}]{Yuan2021BARTScoreEG}
Weizhe Yuan, Graham Neubig, and Pengfei Liu. 2021.
\newblock \href {https://api.semanticscholar.org/CorpusID:235593404} {Bartscore: Evaluating generated text as text generation}.
\newblock \emph{ArXiv}, abs/2106.11520.

\bibitem[{Zhang et~al.(2021)Zhang, Yates, and Lin}]{Zhang2021ComparingSA}
Xinyu~Crystina Zhang, Andrew Yates, and Jimmy~J. Lin. 2021.
\newblock \href {https://api.semanticscholar.org/CorpusID:232423054} {Comparing score aggregation approaches for document retrieval with pretrained transformers}.
\newblock In \emph{European Conference on Information Retrieval}.

\bibitem[{Zheng et~al.(2023)Zheng, Chiang, Sheng, Zhuang, Wu, Zhuang, Lin, Li, Li, Xing, Zhang, Gonzalez, and Stoica}]{zheng2023judging}
Lianmin Zheng, Wei-Lin Chiang, Ying Sheng, Siyuan Zhuang, Zhanghao Wu, Yonghao Zhuang, Zi~Lin, Zhuohan Li, Dacheng Li, Eric.~P Xing, Hao Zhang, Joseph~E. Gonzalez, and Ion Stoica. 2023.
\newblock \href {https://arxiv.org/abs/2306.05685} {Judging llm-as-a-judge with mt-bench and chatbot arena}.
\newblock \emph{Preprint}, arXiv:2306.05685.

\end{thebibliography}

\appendix

\section{Compute Infrastructure}
\label{app:compute}
We run our experiments on 24GB Nvidia Quadro RTX 6000 GPUs for retrieval and 80GB Nvidia A100s for RAG experiments. Retrieval indexing takes around 60 hours for a BERT-base sized model on one GPU. For RAG experiments, they take approximately 2 hours per model on one GPU.

\begin{figure*}
    \centering
    \resizebox{\columnwidth}{!}{%
    \includegraphics{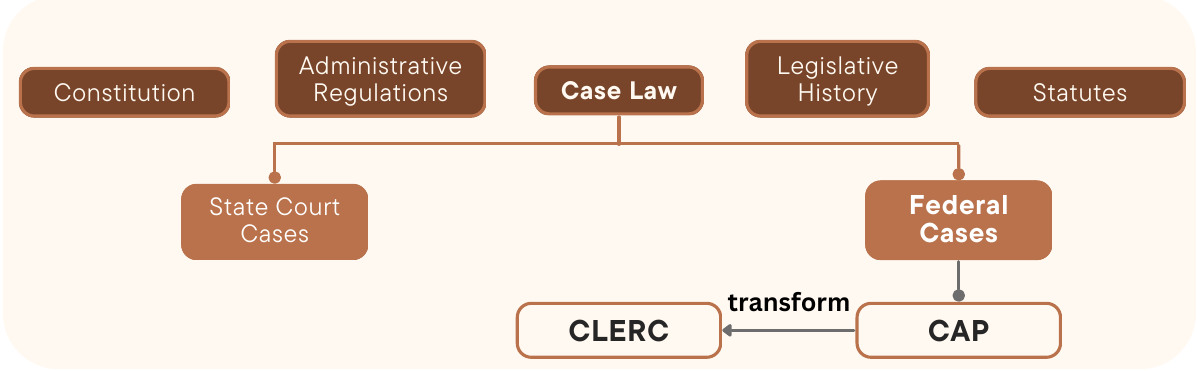}%
    }
    \caption{A taxonomy of U.S. legal data. \clercs is a subset of federal cases, which are a subset of case law. \dataset{} contains millions of documents and has much room for improvement.}
    \label{fig:data-taxonomy}
\end{figure*}
\section{U.S. Legal Data Taxonomy}
Figure~\ref{fig:data-taxonomy} shows a taxonomy of U.S. legal data. We use that data from CAP, which provides U.S. federal cases. Note that U.S. legal text can vary widely according to the type, however, we use federal case law due to its widespread usage and availability.

\section{Prompt for Argument Generation}
\label{app:arg-gen-prompt}

 ~\\~
\texttt{
\textbf{User prompt}: \\
Here are some reference articles for legal cases: \\
\# Reference case \{case\_key\_1\} \\
\{text of cited case 1\} \\
\# Reference case \{case\_key\_2\} \\
\{text of cited case 2\} \\
$\dots$ \\
\# Reference case \{case\_key\_N\} \\
\{text of cited case N\} \\\\
Here is the text I've written so far: \\
\# Paragrah \\
\{previous\_text\} \\\\
Continue to write it following the style of my writeup. Your answer contains 100 to 400 words. You must explicitly use the reference cases and mention their reference ids, i.e. \{case\_key\_1\}, \{case\_key\_2\} \dots \{case\_key\_N\}. Wrap your answer with <answer></answer>. Make your answer concise and avoid redundant languages. \\
}

\texttt{\{previous\_text\}} refers to the previous context before the reference paragraph, i.e. $(p_1, p_2, \dots, p_{t-1})$ defined in Section \ref{exp:arg-gen}. 
\texttt{\{text cited case i\}} refers to the reference articles, i.e. $R_t$ defined in Section \ref{exp:arg-gen}, and \texttt{\{case\_key\_i\}} are the associated citation keys.
 
 For the ablation study, we experiment prompting LLMs without explicitly providing the text of cited cases with the following prompt:
\\~\\

\texttt{\textbf{User prompt}: \\
Here is the text I've written so far: \\
\# Paragrah \\
\{previous\_text\} \\\\
Continue to write it following the style of my writeup. Your answer contains 100 to 400 words. Wrap your answer with <answer></answer>. Make your answer concise and avoid redundant languages. 
}\\

\section{Additional Experimental Results}
\begin{figure*}[t]
    \centering
    \fbox{
    \begin{minipage}{38.5em}
    \footnotesize
    \underline{Types of alterations in direct quote:}\\
    \textbf{Contextual clarification: }\\
    Query: \text{``A reading of the entire [Wage] Act clearly shows that the purpose of the Act is to assist...''}\\
    Original: \text{``A reading of the entire Act clearly shows that the purpose of the Act is to assist...''}\\
    
    \textbf{Rephrases: }\\
    Query: ``punishable for a term of imprisonment exceeding one year.'' \\
    Original: ``punishable by imprisonment for more than one year.'' \\\\
    \textbf{Punctuation changes:}\\
    Query: ``an account of the time, place, and specific content of the false representations as well as the identities of the parties to the misrepresentations.''\\
    Original: an account ``of the time, place, and specific content of the false representations as well as the identities of the parties to the misrepresentations.''
    \end{minipage}
    }
    \caption{Direct quote may experience different types of alterations and degrade retrieval performance.}
    \vspace{-2mm}
    \label{fig:egdq}
\end{figure*}

\subsection{Parse Direct Quotes and Filter Queries}
\label{app:parse}
We design a simple heuristics program to parse direct quotes. CAP uses special double quotation marks for all direct quotes, which have unicode values of 8220 and 8221. We extract all texts quoted by these special quotation marks, and we pair them with the nearest case citation. 

To locate and filter the central citation sentences in the queries, we design an involved RegEx program. The high-level overview is that it will find the start of the citation sentence, which is usually some Entity A v. Entity B, and we recognize the start by referring to the metadata of CAP. There are other variant formats to start a citation sentence, such as \textit{Id.}, \textit{See.}, and \textit{In re}, and we also accomodate these formats in the filter program  \citep{bluebook}. We find the end of the citation by locating some notable punctuations, such as closing parenthesis, semicolon, etc. In the event that we find outlier citation sentences, we would report them as failure cases. The average failure rate is around 11\%.

For more specific details, please refer to our code and repo.

\subsection{Retrieve with Direct Quote}
Around 33\% of the time, we find that case documents cite other cases in the form of direct quotations. We parse such direct quotes from case documents and used them as queries for retrieval, expecting a near-perfect accuracy. However, conducting exact matches with \texttt{grep}, we only retrieve 16\% of the direct quotes. We discover that at least 25.4\% of the direct quotes experience subtle alterations with the use of square brackets (see Figure \ref{fig:egdq}), which partly causes the low recall. 
\label{app:dq}

\begin{table*}[t]
\centering
\small
\begin{tabular}{@{}lrrrrr@{}}
\toprule
        & Recall@5 $\uparrow$ & Recall@10 $\uparrow$ & Recall@100 $\uparrow$ & Recall@1000 $\uparrow$ & nDCG@10 $\uparrow$\\ \midrule
BM25    & 18.8     & 25.2      & \textbf{45.7}       & \textbf{66.7}        & 15.8    \\
5-gram  & 30.3     & \textbf{36.4}      & 44.7       & 44.8        & \textbf{23.0}    \\
12-gram & \textbf{31.1}     & 35.7      & 40.1       & 40.1        & 22.8    \\ \bottomrule
\end{tabular}
\caption{Retrieval experiment results using direct quote as query. All numbers are in percentage.}
\label{tab:dq-res}
\end{table*}

\begin{table*}[ht]
\centering
\small
\begin{tabular}{lllllr}
\toprule
\multirow{2}{*}{Type} & \multirow{2}{*}{Params} & \multirow{2}{*}{Length} & \multirow{2}{*}{Model} & \multicolumn{2}{c}{Metrics} \\
\cmidrule{5-6}
 &  &  &  & R@10 & nDCG@10 \\
\midrule
Bi-Encoders &110M & 512 &  E5 &\textbf{8.4}  &\textbf{3.67}   \\
\midrule
\multirow{3}{*}{Cross-Encoders} 
 & 7B & 512 & E5 + RankLlama &4.7   &2.14   \\
 & 3B & 512 & E5 + MonoT5 &4.2    &2.00   \\
 & 7B & 512 & E5 + FollowIR &5.1  &2.14  \\
\bottomrule
\end{tabular}
\caption{Retrieval results with Cross-Encoders. All numbers are in percentage}
\label{table:cross_encoders}
\end{table*}

Therefore, we further conduct retrieval experiments of direct quotes with BM25 and n-gram match, in the hope that these ``soft'' term matching methods could overcome subtle alterations and yield a higher recall. Results in Table \ref{tab:dq-res} show that the highest Recall@1000 is only 66.7\%, which is far from perfect. 

By examining the case texts, we attribute this result to mainly three reasons. First, direct quote, when being cited, commonly experiences slight alterations in punctuation, wording, and pronouns for clarification purposes or due to human errors (see Figure \ref{fig:egdq}). 

Second, when retrieving a direct quote that consists of common words or widely used legal terms, soft term matching methods could retrieve multiple false positive documents that contain multiple occurrences of these common terms and push down the ranking of the actual positive document. 

Third, direct quotes often appear in contexts of various neighboring case citations that point to a set of candidate documents. Unless investigating and searching all candidate documents, it is not apparent which direct quote corresponds to which case. While we develop a program (see Section \ref{app:dq}) to determine the correspondence based on the proximity of direct quotes to citations, this correspondence determination is imperfect and remains an open question to be addressed in future works.

\subsection{Retrieve with Cross-Encoders}
\label{app:cross-encoders}

We conduct additional retrieval experiments with the same data and metrics setup in Section \ref{exp:retrieval} using multiple Cross-Encoders in Table \ref{table:cross_encoders}. We rerank the top-100 outputs of E5 \citep{wang2023improving} with RankLlama \citep{Ma2023FineTuningLF}, MonoT5-3B \citep{Nogueira2020DocumentRW}, and FollowIR \citep{Weller2024FollowIREA}. We discover that reranking with cross-encoders degrades the retrieval performance. A possible explanation is that due to domain mismatch, rerankers are unable to rerank long-context legal data that have distinct styles and length from the data rerankers were trained on. 

\begin{figure*}[t]
    \centering
    \fbox{
    \begin{minipage}{40em}
    \footnotesize

\textbf{Gold paragraph:}

A pro se complaint, ``however unlawfully pleaded,'' is held ``to less stringent standards than formal pleadings drafted by lawyers.'' Hughes v. Rowe, \textit{\textcolor{darkgreen}{449 U.S. 5, 9, 101 S.Ct. 173, 66 L.Ed.2d 163 (1980)}}. The court's role is to ensure that claims of pro se litigants are given ``fair and meaningful consideration.'' Matzker v. Herr, \textit{\textcolor{darkgreen}{748 F.2d 1142, 1146 (7th Cir.1984)}}. Accordingly, pro se complaints must be liberally construed. Estelle v. Gamble, \textit{\textcolor{darkgreen}{429 U.S. 97, 106, 97 S.Ct. 285, 50 L.Ed.2d 251 (1976)}}. The complaint need not specify the correct legal theory, nor point to the correct statute in order to survive a motion to dismiss. Bartholet v. Reishauer A.G. (Zurich), \textcolor{brown}{\texttt{953 F.2d 1073, 1078 (7th Cir.1992)}}. Finally, a district court may dismiss a complaint only if ``it appears beyond doubt that the plaintiff can prove no set of facts in support of his claim which would entitle him to relief.'' Hughes, supra, 449 U.S. at 10, 101 S.Ct. 173.\\
\rule{\textwidth}{0.4pt}

\textbf{GPT-4o generation:}

Liberal Construction of Pro Se Complaints
\\

The City contends that Mr. Cleaves' complaint should be dismissed because it fails to state a claim upon which relief can be granted. However, it is well-established that complaints filed by pro se litigants must be held to less stringent standards than those drafted by lawyers. As the Supreme Court noted in Haines v. Kerner, “however inartfully pleaded,” a pro se complaint must be liberally construed and should not be dismissed unless it appears beyond doubt that the plaintiff can prove no set of facts in support of his claim which would entitle him to relief (\underline{\textcolor{red4cb}{404 U.S. 519, 520-21 (1972)}}).
\\

Similarly, in Hughes v. Rowe, the Court reiterated this principle, emphasizing that the district court’s role is to ensure that pro se litigants' claims are given fair and meaningful consideration (\goodtext{449 U.S. 5, 9-10}, \badtext{101 S.Ct. 173, 175, 66 L.Ed.2d 163 (1980)}). Courts must determine whether the plaintiff should be allowed to present evidence on his claims before deciding on a motion to dismiss (\goodtext{748 F.2d 1142}).
\\

Additionally, considering the liberal standards applied to such cases as set forth in Estelle v. Gamble (\textcolor{darkgreen}{\textit{429 U.S. 97, 106 (1976)}}), and in light of the notice-pleading standard of the Federal Rules of Civil Procedure (Rule 8(a)), Mr. Cleaves' allegations, though detailed, sufficiently provide the City with notice of the claims and underlying facts.
\\

Therefore, the City’s motion to dismiss Mr. Cleaves' pro se complaint is denied as it fails to demonstrate that there are no conceivable facts under which Mr. Cleaves could prevail.\\
\rule{\textwidth}{0.4pt}

\textbf{Reference texts:}\\
\textbf{449 U.S. 5}

purports to justify or explain the segregation of petitioner for two days in advance of the disciplinary hearing. II Petitioner's complaint, like most prisoner complaints filed in the Northern District of Illinois, was not prepared by counsel. It is settled law that the allegations of such a complaint, “however inartfully pleaded'' are held “to less stringent standards than formal pleadings drafted by lawyers ....'' Haines v. Kerner, \badtext{404 U. S. 519, 520 (1972)}. See also Maclin v. Paulson, 627 F. 2d 83, 86 (CA7 1980); French v. Heyne, 547 F. 2d 994, 996 (CA7 1976). $\cdots$ (omitted for brevity) $\cdots$

\textbf{748 F.2d 1142}

magistrate erred in applying the deliberate indifference standard in dismissing this complaint. III. Our inquiry turns, therefore, to the question of whether Matzker stated a cause of action when judged under due process standards. A complaint drafted by a pro se litigant “however inartfully pleaded,” is held “to less stringent standards than formal pleadings drafted by lawyers.” Hughes v. Rowe, 449 U.S. 5, 9, \badtext{101 S.Ct. 173, 175, 66 L.Ed.2d 163 (1980)}. The district court’s role is to ensure that the claims of pro se litigants are given “fair and meaningful consideration.” $\cdots$ (omitted for brevity) $\cdots$
\\
\textbf{429 U.S. 97}

to evidence deliberate indifference to serious medical needs. It is only such indifference that can offend “evolving standards of decency” in violation of the Eighth Amendment. Ill Against this backdrop, we now consider whether respondent’s complaint states a cognizable § 1983 claim. The handwritten pro se document is to be liberally construed. $\cdots$ (omitted for brevity) $\cdots$

\textbf{953 F.2d 1073}

it done so, ERISA still would govern unless the plan were “unfunded.” The exemption recognizes that a suit concerning an unfunded plan is one directly against the employer’s assets, and as ERISA leaves excess benefit plans substantively unregulated there is no reason to oust state law. $\cdots$ (omitted for brevity) $\cdots$\\
\end{minipage}
} 
    \caption{Example GPT-4o generation that exhibits hallucination and low citation precision and recall. The input of generation is all referenced texts concatenated with all paragraphs before the gold paragraph in the original case document. The generation is supposed to generate all citations in the gold paragraph. \goodtext{Green italicized texts} are citations successfully generated, \misstext{yellow texts in teletype fonts} are citation missed in generation, and
    \badtext{red underlined texts} are hallucinated citations. GPT-4o should cite all referenced cases (449 U.S. 5, 748 F.2d 1142, 429 U.S. 97, 953 F.2d 1073). Instead, since we provide the actual texts of referenced cases, \textbf{GPT-4o hallucinates citations from citations in referenced cases}, which we underline with red color. In summary, the generation achieves the following results on citation metrics: $CP=3/5$, $CR=3/4$, $CFP=2/5$.}
    \vspace{-2mm}
    \label{fig:eghalu}
\end{figure*}

\section{Qualitative Example}
\label{app:addeg}
Please see Figure \ref{fig:eghalu} for an example of problematic legal analysis generated by GPT-4o, which exhibits hallucination and achieves low citation precision and recall. We discover that citations from the cited cases are a possible source of hallucination, meaning that the model confuses the cases they should cite with the citations \underline{in} those cases. It is possible to mitigate this with more refined prompt engineering, which we will leave to future works.

\end{document}